\documentclass[10pt,twocolumn,letterpaper]{article}

\usepackage[pagenumbers]{cvpr} 

\usepackage{cuted} 

\usepackage{caption} 

\usepackage{microtype}

\renewcommand{\paragraph}[1]{\vspace{.5em}\noindent\textbf{#1.}}

\usepackage{xspace}

\usepackage{algorithm}
\usepackage{algpseudocode}
\usepackage[breakable,skins]{tcolorbox}

\newtcolorbox{promptbox}[1][]{%
  enhanced, breakable,
  colback=gray!5, colframe=gray!55,
  boxrule=0.4pt, arc=2pt,
  left=6pt, right=6pt, top=4pt, bottom=4pt,
  fonttitle=\bfseries\small\sffamily,
  fontupper=\ttfamily\small,
  coltitle=black, colbacktitle=gray!20,
  attach boxed title to top left={xshift=6pt, yshift=-2pt},
  boxed title style={colframe=gray!55, boxrule=0.4pt, arc=2pt},
  #1%
}

\newcommand{\metric}{LPB\xspace}           
\newcommand{\method}{Seq2SHAP\xspace}      

\definecolor{cvprblue}{rgb}{0.21,0.49,0.74}
\usepackage[pagebackref,breaklinks,colorlinks,allcolors=cvprblue]{hyperref}

\title{Do VLMs Answer Without Looking? \\Quantifying Language Prior Bias via Shapley Attribution}

\author{Nicholas Mesa-Cucalon$^{1}$\thanks{Equal contribution.} \quad Yinong Oliver Wang$^{1}$\footnotemark[1] \quad Ting-Yao Hu$^{2}$ \quad Raviteja Vemulapalli$^{2}$\\
Oncel Tuzel$^{2}$ \quad Nicholas Apostoloff$^{2}$ \quad Jerremy Holland$^{2}$ \quad Fernando De la Torre$^{1}$\\[0.5em]
$^{1}$Carnegie Mellon University \qquad $^{2}$Apple
}

\begin{document}
\maketitle


\begin{strip}
  \centering
  \vspace{-14mm}
  \includegraphics[width=\linewidth]{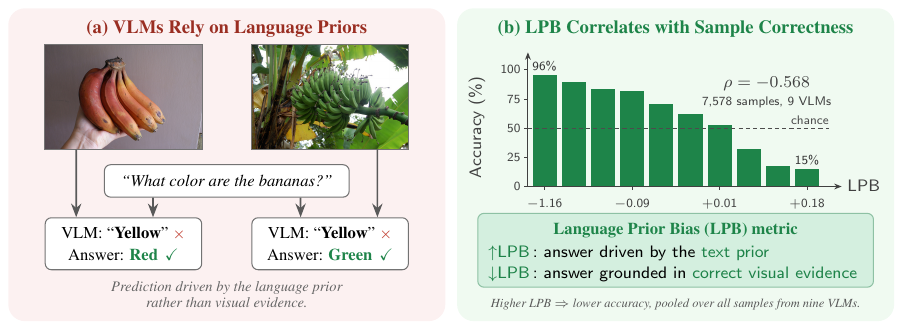}
  \captionsetup{hypcap=false}
  \captionof{figure}{\textbf{VLMs can ``answer without looking'' (i.e., rely on language rather than visual input), and our metric \metric measures it.}
  \textbf{(a)}~When asked the same question on visibly non-yellow bananas, a VLM could return ``yellow'' regardless of the image, as a result of the language prior bias.
  \textbf{(b)}~Our token-level metric \metric quantifies such bias. On VLind-Bench-LP, pooling all samples from nine VLMs into ten equal-size \metric bins, accuracy falls from 96\% in the lowest bin to 15\% in the highest (Spearman $\rho=-0.568$). Tick labels give bin means.}
  \captionsetup{hypcap=true}
  \label{fig:teaser}
\end{strip}

\begin{abstract}
Vision-Language Models (VLMs), despite strong benchmark performance, suffer from an implicit yet critical issue: VLMs can shortcut to answers dictated by language prior statistics rather than the actual visual evidence, a phenomenon known as \textbf{Language Prior Bias} that leads to answers that overlook the visual input and undermines the trustworthiness of VLMs.
Existing diagnostic methods detect this bias at the benchmark or modality level, but they do not quantify, token by token, how much each input pulls a prediction toward the language prior.
We propose \textbf{\metric}, a Shapley-value-based metric that decomposes each VLM prediction into per-token contributions and scores how much it is driven by text rather than the image.
To make \metric computation practical, we also introduce \textbf{\method}, an amortized Shapley estimator for autoregressive VLMs, which 
is $187\times$ faster than KernelSHAP with little loss in accuracy. We show that \metric negatively correlates with correctness on VLind-Bench-LP at the sample level (Spearman $\rho=-0.568$ pooled across nine VLMs) and signals prior-driven errors (AUROC $0.835$ pooled across nine VLMs). Beyond diagnosing, our metric is also actionable as a fine-tuning objective (+4.5\% average accuracy gain across eight VLMs) and as inference-time guidance to improve VLMs.
Together, \metric and \method make language prior bias measurable for every prediction, without labels, a step toward trustworthy VLMs that answer from what they see, not from their language prior.
\end{abstract}

\section{Introduction}
\label{sec:introduction}

Vision-Language Models (VLMs) have achieved impressive performance across a wide range of multimodal tasks, from visual question answering (VQA) to image captioning and video understanding~\cite{radford2021clip, alayrac2022flamingo, liu2024llava, dai2023instructblip, wang2024qwen2}. Yet a growing body of evidence reveals a fundamental weakness: VLMs are prone to overlooking visual input and shortcut to language priors, which are statistical regularities absorbed from large-scale text corpora during pretraining~\cite{goyal2017making, lee2024vlindbench, leng2024vcd, tong2024mmvp}. A VLM may answer ``Is there a dog next to the person?'' correctly not because it perceives the dog, but because dogs frequently co-occur with human in its training data. This phenomenon, known as \emph{Language Prior Bias}, yields visually ungrounded answers (\cref{fig:teaser}a), undermining VLM reliability.

Prior work addresses language prior bias through balanced datasets~\cite{goyal2017making}, benchmarks~\cite{lee2024vlindbench}, and decoding-time mitigation~\cite{leng2024vcd}, treating its \emph{symptoms} without measuring how much each input contributes to a prediction. Existing diagnostics measure bias by its outcome, as accuracy gaps without the image~\cite{chen2024mmstar}, or by attention mass pooled per modality~\cite{wu2025language}; neither attributes a prediction to its inputs. 
We instead quantify language prior bias with an interpretable, actionable per-sample metric, obtained by attributing each prediction to its individual image and text tokens.

Shapley values from cooperative game theory~\cite{shapley1953value} provide a natural framework for fine-grained attribution, treating each token as a player and the model prediction as the payout.
However, existing Shapley analyses of VLMs aggregate unsigned contributions per modality via Monte-Carlo sampling~\cite{parcalabescu2023mmshap,parcalabescu2025decoders,pollard2025videoworththousandwords}.
Scaling to token-level attribution in modern VLMs also faces a computational barrier, as one image can span over a thousand tokens~\cite{wang2024qwen2}. 
KernelSHAP~\cite{lundberg2017unified} requires thousands of forward passes per instance, and amortized estimators such as FastSHAP~\cite{jethani2022fastshap} assume fixed-length inputs instead of variable-length multimodal sequences.

We overcome these limitations with finer-grained attribution and efficient Shapley value predictor. First, we formulate a token-level cooperative game whose Shapley values aggregate into a sample-level score, \textbf{\metric}, directly quantifying how much a VLM relies on the language prior. Second, we introduce \textbf{\method}, an amortized Shapley method for autoregressive multimodal models, reducing attribution from thousands of forward passes to a constant cost: two VLM inferences plus one pass through a lightweight explainer.

We show that \method closely approximates KernelSHAP with a $187\times$ speedup. Moreover, \metric correlates negatively with correctness at the sample level on VLind-Bench-LP~\cite{lee2024vlindbench}: pooled over the nine VLMs we evaluate, the sample-wise Spearman correlation between \metric and prediction correctness is significantly negative at $-0.568$.
This trend is consistent with the hypothesis that language prior bias is a performance impediment.
We further show the metric is actionable: fine-tuning with \metric as a regularizer improves model accuracy by $4.5$ points over eight VLMs, and prompt refinement guided by \metric corrects $30.2\%$ of originally wrong answers, against $1.9\%$ for a fixed neutral prefix.
In summary, our contributions are:
\begin{enumerate}
  \item \metric, a Shapley-based metric for quantifying language prior bias in VLMs;
  \item \method, an amortized Shapley method for autoregressive multimodal models; and
  \item demonstration that \metric is both diagnostic and actionable as a training objective and guides prompt refinement.
\end{enumerate}


\section{Related Work}
\label{sec:related_work}

\paragraph{Language prior and modality bias in VLMs}
Language shortcuts over visual evidence have been examined since early Visual Question Answering (VQA) studies. VQA v2.0~\cite{goyal2017making} and VQA-CP~\cite{agrawal2018vqacp} exposed answer-prior shortcuts via balanced pairs and out-of-distribution splits; RUBi~\cite{cadene2019rubi} factored out a question-only branch, LPF~\cite{liang2021lpf} re-weighted samples requiring visual reasoning, and GQA-OOD~\cite{kervadec2021roses} exposed residual shortcuts on rare concepts. The problem persists at scale: VLind-Bench~\cite{lee2024vlindbench} disentangles language priors from confounders, ViLP~\cite{luo2025probing} shows VLMs fall short on visually grounded questions, and mechanistic studies localize a ``Visual Integration Point''~\cite{long2025understanding} and an information-transfer pathway~\cite{basu2024understanding}, with related efforts in video QA~\cite{park2024modality,xiao2021nextqa,chakraborty2026notanother}. Recent work quantifies modality reliance via benchmark leakage~\cite{chen2024mmstar} and attention-based text-dominance indices~\cite{wu2025language}, but only at the modality level.

\paragraph{Shapley-based multimodal explanations}
Shapley values~\cite{shapley1953value} were first brought to ML attribution through Monte Carlo sampling~\cite{strumbelj2014explaining} and later through the regression-based KernelSHAP~\cite{lundberg2017unified}, while SAGE~\cite{covert2020sage} extended them from individual predictions to global feature importance. MM-SHAP~\cite{parcalabescu2023mmshap} measures modality-level contributions in VL encoders, and \citet{parcalabescu2025decoders} extend it to autoregressive VLMs, but its unsigned scores ignore whether an input pulls toward or away from the language prior, and its sampling cost limits evaluation to small subsets. Dr.\ SHAP-AV~\cite{cappellazzo2026drshapav} applies modality-level Shapley attribution to audio-visual speech recognition. MultiSHAP~\cite{wang2025multishap} adds Shapley interactions over patches and tokens but is expensive due to pairwise masking. We instead provide amortized token-level attribution for autoregressive VLMs.

\paragraph{Amortized and efficient Shapley methods}
FastSHAP~\cite{jethani2022fastshap} amortizes Shapley computation through a learned single-pass explainer, and FastSHAP-KL~\cite{jethani2023label} addresses its label leakage. HarsanyiNet~\cite{chen2023harsanyinet} computes exact Shapley values in one pass but requires architectural changes. For language models, \citet{yang2023amortized} achieve a 60$\times$ speedup on text classification, and TextGenSHAP~\cite{enouen2024textgenshap} reduces LM explanation from hours to minutes. None of these target autoregressive VLMs with their variable-length cross-modal inputs sequences; \method is the first amortized Shapley method for this setting.

\paragraph{VLM hallucination and benchmarks}
Language prior bias often surfaces as hallucination, a link articulated early by CHAIR~\cite{rohrbach2018chair}. POPE~\cite{li2023pope} polls for object hallucination; HallusionBench~\cite{guan2024hallusionbench} separates language hallucination from visual illusion; Winoground~\cite{thrush2022winoground} probes compositionality with shared-vocabulary image-caption pairs; V-FAT~\cite{wang2026vfat} pits atypical images and misleading instructions against language priors; and MVP~\cite{krojer2025mvp} pairs near-identical videos with opposite answers. Mitigations include contrastive decoding (VCD~\cite{leng2024vcd}, ICD~\cite{wang2024icd}), contrastive training (S-VCO~\cite{wu2025svco}), and post-hoc correction (LURE~\cite{zhou2024lure}). These works address the symptoms of bias; our metric complements them by quantifying how much each token drives a prediction.

\section{Background}
\label{sec:background}

\paragraph{Shapley values}
Consider a cooperative game with a set of players $\mathcal{N} = \{1, \ldots, N\}$ and a value function $v : 2^{\mathcal{N}} \to \mathbb{R}$ satisfying $v(\emptyset) = 0$. The Shapley value $\phi_i$ of player $i$ is the unique attribution satisfying efficiency, symmetry, linearity, and the dummy axiom~\cite{shapley1953value}:
\begin{equation}
\label{eq:shapley}
\phi_i = \sum_{S \subseteq \mathcal{N} \setminus \{i\}} \frac{|S|!\,(N - |S| - 1)!}{N!} \bigl[v(S \cup \{i\}) - v(S)\bigr].
\end{equation}
The efficiency property guarantees that $\sum_{i \in \mathcal{N}} \phi_i = v(\mathcal{N}) - v(\emptyset)$, i.e., the attributions fully decompose the value of the grand coalition $v(\mathcal{N})$. In our setting, players are question tokens and visual tokens, and value function $v$ captures the model's output behavior (\cref{sec:problem_formulation}), so Shapley values provide a principled decomposition of each model prediction into per-player contributions.

\paragraph{KernelSHAP}
Exact computation of~\cref{eq:shapley} requires $O(2^N)$ evaluations of $v$, which is intractable for inputs with thousands of tokens. KernelSHAP~\cite{lundberg2017unified} instead recovers Shapley values by regression: it treats the attributions as coefficients of an additive model of the value function, $v(S) - v(\emptyset) \approx \sum_{i \in S} \phi_i$, and fits them by weighted least squares over coalitions drawn from the Shapley kernel
\begin{equation}
\label{eq:shapley_kernel}
\pi(S) \propto \frac{N-1}{\binom{N}{|S|}\,|S|\,(N-|S|)}, \quad 0 < |S| < N,
\end{equation}
which concentrates samples on very small and very large coalitions, where a single player's effect is most visible:
\begin{equation}
\label{eq:kernelshap}
\hat{\boldsymbol{\phi}} = \arg\min_{\boldsymbol{\phi}} \; \mathbb{E}_{S \sim \pi} \Bigl[ \bigl(v(S) - v(\emptyset) - \textstyle\sum_{i \in S} \phi_i \bigr)^2 \Bigr],
\end{equation}
subject to efficiency, $\sum_{i \in \mathcal{N}} \phi_i = v(\mathcal{N}) - v(\emptyset)$. \citet{lundberg2017unified} show that the solution is then exactly the Shapley value. In practice, the expectation is approximated with sampled coalitions, each costing one forward pass, and an accurate estimate can need thousands per instance, which is impractical for VLMs.

\paragraph{FastSHAP}
FastSHAP~\cite{jethani2022fastshap} amortizes this cost by training an \emph{explainer} network $\theta$ to predict $\hat{\phi}_i^{(\theta)}(\mathbf{x})$ for any input $\mathbf{x}$ in a single forward pass. For a model $f$ and coalition $S \subseteq \mathcal{N}$, $\mathbf{x}_S$ denotes $\mathbf{x}$ with the players outside $S$ removed by an \emph{imputer} (\cref{sec:seq2shap}); with a \emph{link} $g$ mapping raw model outputs to the game's value space, $v(S) = g\bigl(f(\mathbf{x}_S)\bigr)$. The explainer replaces the per-instance solution of~\cref{eq:kernelshap} and is trained on the same objective in expectation over inputs:
\begin{equation}
\label{eq:fastshap}
\begin{aligned}
\mathcal{L}(\theta) = {} & \mathbb{E}_{\mathbf{x}}\, \mathbb{E}_{S \sim \pi} \Bigl[ \bigl(v(S) - v(\emptyset) - \textstyle\sum_{i \in S} \hat{\phi}_i^{(\theta)}(\mathbf{x}) \bigr)^2 \Bigr], \\
\text{s.t.} \quad & \textstyle\sum_{i \in \mathcal{N}} \hat{\phi}_i^{(\theta)}(\mathbf{x}) = v(\mathcal{N}) - v(\emptyset).
\end{aligned}
\end{equation}
FastSHAP satisfies the constraint by shifting the explainer outputs, where $v(\emptyset)$ is the value with every player removed.

\section{Method}
\label{sec:method}

\begin{figure*}[t]
  \centering
  \includegraphics[width=\linewidth]{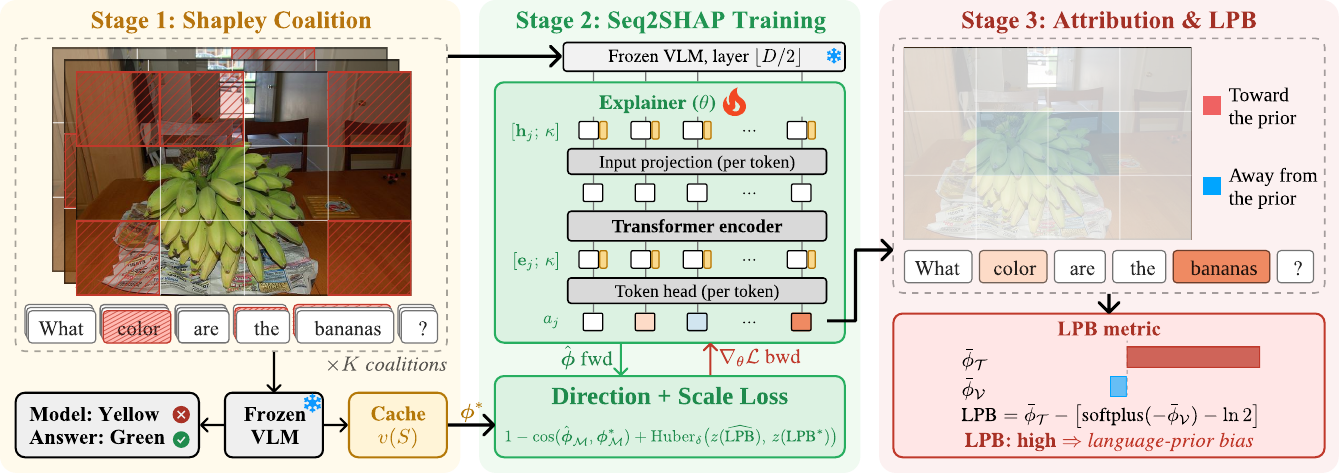}
  \vspace{-3mm}
  \caption{\textbf{Method overview.} Stage~1 \textbf{coalition cache} (\cref{sec:seq2shap}): for each data point, we sample $K$ coalitions and run the frozen VLM through an attention-mask imputer to cache their values; striped players are masked out. Stage~2 \textbf{\method explainer} (\cref{sec:seq2shap}): each token's middle-layer VLM hidden state $\mathbf{h}_j$ (white input blocks), with $\kappa$ (small yellow blocks, the divergence of the full-input prediction from the language prior) appended, passes through the input projection, Transformer encoders that mix the tokens, and the token head where $\kappa$ is appended again, outputting one score $a_j$ per token, averaged within each player into $\hat{\phi}_i$; the explainer is trained offline on the estimated Shapley target $\boldsymbol{\phi}^{*}$. Stage~3 \textbf{\metric metric} (\cref{sec:lpb_metric}): the attributions are averaged per modality and combined into the \metric score; red and blue mark a pull toward and away from the language prior.}
  \label{fig:method_overview}
\end{figure*}

We first define the cooperative game where Shapley values attribute language prior bias to individual tokens (\cref{sec:problem_formulation}). With these values, we build the token-level quantification metric, \metric (\cref{sec:lpb_metric}). We then introduce \method, which estimates these Shapley values with a lightweight explainer (\cref{sec:seq2shap}). \Cref{fig:method_overview} shows the full pipeline.

\subsection{Problem Formulation}
\label{sec:problem_formulation}

We consider VLMs solving multiple-choice tasks: given an image and a user-text prompt that comprises a question, together with $C$ answer options and an instruction, the model predicts a distribution over the option set and emits a single answer-position (first output) token. \cref{app:extension} shows the natural extension to open-ended generation.

\paragraph{Player definition} After tokenization, the user input $\mathbf{x}$ consists of visual patch tokens encoded from the image and question tokens drawn from the question. Only the visual tokens and the question tokens are treated as attributable players.
The trailing answer instruction and chat-template scaffolding (role markers, system tokens) are held fixed across all coalitions, which keeps text-side attribution focused on the semantic content of the question. The full prompt templates and their player/scaffolding partitions are listed in~\cref{app:prompts}.
We denote the resulting player set $\mathcal{N} = \mathcal{T} \cup \mathcal{V}$, with $\mathcal{T}$ indexing text players and $\mathcal{V}$ indexing visual players, and write $N = |\mathcal{N}|$, $N_{\mathcal{T}} = |\mathcal{T}|$, and $N_{\mathcal{V}} = |\mathcal{V}|$. To keep the cooperative game balanced between modalities, in practice, each visual player is a spatial super-token formed by grouping visual patches, while each textual player remains a single token; details are deferred to~\cref{app:implementation}. 

\paragraph{Value function}
To quantify language prior bias, we measure how much each player pulls the model toward its own language prior. Let $p_{\text{LM}}$ denote the model's answer distribution given the text alone, and let $p_S$ denote its answer distribution on $\mathbf{x}_S$, the input in which only the players in $S \subseteq \mathcal{N}$ are present; $p_{\mathcal{N}}$ is the full-input distribution and $p_\emptyset$ is the null-coalition distribution. All distributions are computed by renormalizing the model's full vocabulary distribution over the $C$ option tokens at the answer position. Following the link $g$ of FastSHAP (\cref{sec:background}), we define $g(\cdot) = \mathrm{KL}\left(p_{\text{LM}}\|\cdot\right)$. Then we define our value function:
\begin{equation}
\label{eq:value_function}
v(S) = \mathrm{KL}\!\left(p_{\text{LM}} \,\|\, p_\emptyset\right) - \mathrm{KL}\!\left(p_{\text{LM}} \,\|\, p_S\right).
\end{equation}
It follows that $v(\emptyset) = 0$ by construction. A positive $v(S)$ means the coalition $S$ moves the model's output closer to its language prior, and a negative $v(S)$ means it moves the output away from it.
With this design, the game does not require any ground truth label and attributions measure the direction of each player's pull relative to the prior independently of correctness (yet we discuss the extendability to ground the value function based on correctness as well in~\cref{app:extension}). Subtracting the per-sample null anchors every game to its own baseline, preventing null-value drift across samples whose prompt templates differ.

\subsection{Language Prior Bias Metric}
\label{sec:lpb_metric}

While the Shapley values $\phi_i(\mathbf{x})$ of the game in \cref{eq:value_function} provide language prior bias explanation at token-level for a given sample $\mathbf{x}$, we also quantify the extent of bias as a sample-level scalar metric for straightforward evaluation. Thus, we aggregate the fine-grained attribution on the modality level (\cref{fig:method_overview} stage 3). Since the number of text ($N_{\mathcal{T}}$) and vision ($N_{\mathcal{V}}$) players can differ substantially across samples and modalities, we work with per-modality means:
\begin{equation}
\label{eq:mean_shapley}
\bar{\phi}_{\mathcal{T}}(\mathbf{x}) = \frac{1}{N_{\mathcal{T}}} \sum_{i \in \mathcal{T}} \phi_i(\mathbf{x}), \qquad \bar{\phi}_{\mathcal{V}}(\mathbf{x}) = \frac{1}{N_{\mathcal{V}}} \sum_{i \in \mathcal{V}} \phi_i(\mathbf{x}).
\end{equation} A positive $\bar{\phi}_{\mathcal{T}}$ means the text steers the model toward the answer it would give without looking, which is the language-prior reliance we aim to measure. On the vision side, a negative $\bar{\phi}_{\mathcal{V}}$ means the visual input overrides the prior, while a non-negative $\bar{\phi}_{\mathcal{V}}$ (vision neutral or reinforcing the prior) carries no evidence of grounding. To credit only vision that actively overrides the prior, we apply an asymmetric function~\cite{frye2020asymmetric} to the negated vision mean:
\begin{equation}
\label{eq:lpb}
\text{\metric}(\mathbf{x}) \;=\; \bar{\phi}_{\mathcal{T}}(\mathbf{x}) - \bigl[\text{softplus}(-\bar{\phi}_{\mathcal{V}}(\mathbf{x})) - \ln 2\bigr].
\end{equation}
The vision term is a softened ReLU that keeps \metric differentiable, so it can serve as a training signal (\cref{sec:actionable}), and $-\ln 2$ makes the term zero when $\bar{\phi}_{\mathcal{V}} = 0$. When vision overrides the prior ($\bar{\phi}_{\mathcal{V}} \ll 0$), the term approaches $-\bar{\phi}_{\mathcal{V}}$, so $\text{\metric} \approx \bar{\phi}_{\mathcal{T}} + \bar{\phi}_{\mathcal{V}}$ and the sample is marked as low-bias. When vision is neutral or reinforces the prior ($\bar{\phi}_{\mathcal{V}} \geq 0$), the term saturates near zero and $\text{\metric} \approx \bar{\phi}_{\mathcal{T}}$. Note that our scope is measuring over-reliance on text rather than diagnosing why vision fails. (\cref{app:metric_ablations}).
Lower \metric values mean the prediction moved further from the model's text-only answer. Finally, we denote the model-level score as the mean of $\text{\metric}(\mathbf{x})$ over the evaluation set.

\subsection{\method}
\label{sec:seq2shap}

To make token-level attribution tractable for modern VLMs, we introduce \method, which adapts the FastSHAP framework (\cref{sec:background}) to variable-length, multimodal sequences. \method has two stages: coalition cache generation per sample, and explainer training to predict per-token Shapley values for any new sample.

\paragraph{Stage 1: Coalition Preparation}
\cref{fig:method_overview} stage 1 builds the per-sample (coalition, value) pairs that the explainer will later regress on. The mechanism for realizing a coalition $S$ over the frozen VLM is an \emph{attention-mask imputer}: we zero the attention-mask entries at selected positions, effectively masking out these tokens (see \cref{app:imputer} for imputer ablation). For each sample, we draw $K$ coalitions $S_1, \ldots, S_K$ from the Shapley kernel $\pi$ (\cref{eq:shapley_kernel}). For each coalition $S$, we run the frozen VLM with the attention mask zeroed at every token of every player outside $S$, and obtain the answer distribution $p_S$ over the $C$ options. Additionally, we compute the text-only distribution $p_{\text{LM}}$ and the null-coalition distribution $p_\emptyset$ once per sample. See \cref{app:coalition-generation} for the full pseudo-algorithm.

\paragraph{Stage 2: Explainer architecture and training}
The \method explainer predicts one Shapley value per player from two inputs. The first is $\mathbf{h}_j \in \mathbb{R}^{d}$, the frozen VLM's hidden state of token $j$ at the middle layer $\ell = \lfloor D/2 \rfloor$ of its $D$ decoder layers (ablated in \cref{app:seq2shap}), for a sample $\mathbf{x}$ of length $L$. The second is $\kappa = g(p_{\mathcal{N}})$, the divergence of the full-input prediction from the language prior. The architecture (\cref{fig:method_overview} stage 2) comprises:

\begin{itemize}
    \item \textbf{Input projection.} For each token $j$ of the input sequence, we construct the explainer input $[\mathbf{h}_j;\, \kappa]$, which a linear layer projects to the explainer width. Appending $\kappa$ gives every position the sample-level distance to the prior, which sets the scale of the attributions under \cref{eq:value_function}.
    \item \textbf{Transformer encoder.} A Pre-LN~\cite{xiong2020layer} encoder with grouped-query attention (GQA)~\cite{ainslie2023gqa}, SwiGLU~\cite{shazeer2020glu} feed-forward networks, and RMSNorm~\cite{zhang2019rmsnorm}, which outputs one state $\mathbf{e}_j$ per position $j$.
    \item \textbf{Token head.} A linear layer maps each encoder state $\mathbf{e}_j$, with $\kappa$ appended again, to a scalar $a_j$ for token $j$. A player can span several tokens (\eg, a visual super-token), so we average over the tokens of each player, $\hat{\phi}_i = \frac{1}{|\mathcal{J}_i|} \sum_{j \in \mathcal{J}_i} a_j$, and obtain one value per player. As in FastSHAP, the player values are then shifted additively to enforce the efficiency axiom of Shapley values.
\end{itemize}

\textbf{Training target.} FastSHAP (\cref{eq:fastshap}) trains \emph{online}, querying $v$ on fresh coalitions every step, which is affordable only when evaluating $v$ is cheap; for a VLM, every evaluation is a full forward pass. Since Stage~1 already caches $K$ Shapley-kernel coalitions per sample, we instead regress the explainer offline~\cite{yang2023efficient,covert2024stochastic} onto precomputed closed-form KernelSHAP estimates (\cref{app:offline}). With $\Delta = v(\mathcal{N}) = g(p_\emptyset) - g(p_{\mathcal{N}})$ being the explained total, we solve
\begin{equation}
\label{eq:phi_star}
\begin{aligned}
\boldsymbol{\phi}^{*} = &\arg\min_{\boldsymbol{\phi}} \frac{1}{K}\sum_{k=1}^{K}
\bigl(v(S_k)-S_k^\top\boldsymbol{\phi}\bigr)^2 \\
&\text{s.t.} \quad \textstyle\sum_{i\in\mathcal{N}}\phi_i=\Delta,
\end{aligned}
\end{equation}
where $S_k\in\{0,1\}^N$ is the indicator vector of the $k$-th cached coalition and the constraint is the efficiency axiom. These regression targets are noisy but approximately unbiased, and \citet{covert2024stochastic} show that training on such labels is effective. We discuss this in further detail in \cref{app:offline}.

\textbf{Loss.} The explainer is trained with two terms. $\mathcal{L}_{\text{dir}}$ matches the direction of the attribution inside each modality, and $\mathcal{L}_{\text{scale}}$ matches its scale through the value of \metric{} it produces (\cref{sec:lpb_metric}). Our final loss is
\begin{equation}
\label{eq:training_loss}
\begin{aligned}
&\mathcal{L}(\theta) = \mathcal{L}_{\text{dir}} + \mathcal{L}_{\text{scale}}, \quad\quad\quad\text{with} \\
&\mathcal{L}_{\text{dir}} = \tfrac{1}{2} \textstyle\sum_{\mathcal{M} \in \{\mathcal{T}, \mathcal{V}\}} \bigl(1 - \cos(\hat{\boldsymbol{\phi}}_{\mathcal{M}}, \boldsymbol{\phi}^{*}_{\mathcal{M}})\bigr), \\
&\mathcal{L}_{\text{scale}} = \mathrm{Huber}_{\delta}\bigl(z(\widehat{\text{\metric}}),\; z(\text{\metric}^{*})\bigr),
\end{aligned}
\end{equation}
where $\hat{\boldsymbol{\phi}}_{\mathcal{M}}$ and $\boldsymbol{\phi}^{*}_{\mathcal{M}}$ are the predicted and target attributions restricted to the players of modality $\mathcal{M}$. We take the cosine separately per modality since vision tokens often outnumber text tokens; one cosine loss over the whole sequence is dominated by vision and insensitive to errors in the text attributions. However, cosine distance is invariant to scale, making $\mathcal{L}_{\text{dir}}$ blind to magnitude of either modality, and hence to the text-to-vision balance that \metric measures. $\mathcal{L}_{\text{scale}}$ supplies this missing scale, since \metric depends on the attributions through the modality means evaluated on $\hat{\boldsymbol{\phi}}$ and $\boldsymbol{\phi}^{*}$ respectively. $z(u) = (u - \mu)/\sigma$ standardizes with the mean $\mu$ and standard deviation $\sigma$ of $\text{\metric}^{*}$ on the training splits.

\section{Experiments}
\label{sec:experiments}

\subsection{Experimental Setup}
\label{sec:setup}

\paragraph{Models} We evaluate across nine instruction fine-tuned VLMs. Within the Qwen-VL family, we include Qwen2-VL-7B~\cite{wang2024qwen2}, Qwen2.5-VL-7B~\cite{bai2025qwen25vl} and Qwen3-VL-\{2B, 4B, 8B, 32B\}~\cite{bai2025qwen3vl}, where the latter also serves as a scaling sweep. We also include Tarsier-7B~\cite{wang2024tarsier}, InternVL3-8B~\cite{zhu2025internvl3}, and Molmo2-8B~\cite{clark2026molmo2} as cross-architecture baselines.

\paragraph{Datasets} We use three image question-answering datasets. \textbf{VLind-Bench-LP}~\cite{lee2024vlindbench} pairs counter-commonsense images with prior-conflicting statements, designed to isolate the effect of language-prior bias. \textbf{VQAv2}~\cite{goyal2017making} (yes/no subset) adds natural images and questions. The \method explainer is trained on a mix of the two, with a random $10\%$ subset of VLind-Bench-LP held out for validation. We found that this helps improve \method generalize to unseen data.
We also use \textbf{POPE}~\cite{li2023pope}, a yes/no object-hallucination benchmark over MS-COCO, which serves as a held-out set the explainer never sees. VQAv2 is also built on MS-COCO, so we remove every VQAv2 sample whose image appears in POPE.

\paragraph{Metrics} For each sample task, we report: (1) \textbf{Accuracy} on the question-answering tasks above, and (2) \textbf{\metric}, our Shapley-based metric for Language Prior Bias. We correlate \metric with prediction correctness at the sample-level using \textbf{Pearson $r$}, which measures linear association in \metric's native units, and \textbf{Spearman $\rho$}, which assesses monotone relationship. We also report the \textbf{AUROC} of predicting incorrect answers from \metric, with 95\% percentile-bootstrap confidence intervals (CIs) on $421$ VLind-Bench-LP samples.

\paragraph{Implementation details} Prompt templates are shown in \cref{app:prompts}. Details on vision preprocessing, spatial super-tokens, KernelSHAP configuration, and the \method explainer architecture and training recipe are in \cref{app:implementation}.

\begin{figure}[t]
  \centering
  \includegraphics[width=\linewidth]{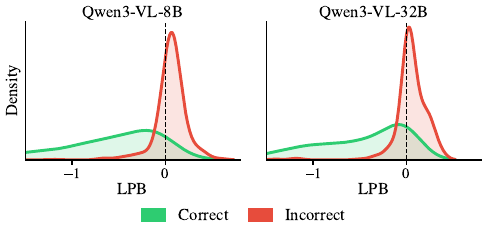}
  \vspace{-6mm}
  \caption{\textbf{Distribution of \metric on correct and incorrect predictions of Qwen3-VL-8B and Qwen3-VL-32B on VLind-Bench-LP.} The correct response density peaks at a lower \metric compared to the incorrect answers, reflecting non-trivial patterns and correlations between \metric and prediction correctness. The dashed line marks \metric$=0$. The x-axis is clipped to $[-1.5, 0.8]$, which hides the left tail of correct samples (4.5\% of all samples).}
  \label{fig:correlation}
  \vspace{-3mm}
\end{figure}

\subsection{Correlation with Performance}
\label{sec:correlation_accuracy}
If \metric measures a meaningful failure mode of visual grounding, then high language-prior reliance should be associated with under-performance.

\paragraph{Sample-wise correlation} Pooling all VLind-Bench-LP samples across nine models ($n{=}7{,}578$; \cref{fig:teaser}b), \metric correlates with prediction correctness at Spearman $\rho = -0.568$ (95\% CI $[-0.583, -0.552]$) and Pearson $r = -0.364$ (95\% CI $[-0.376, -0.351]$), both with $p < 10^{-4}$. \metric also predicts incorrect answers at AUROC $= 0.835$ (95\% CI $[0.826, 0.844]$), with $p < 10^{-4}$ against chance. 
\Cref{fig:correlation} shows this separation for Qwen3-VL-8B and Qwen3-VL-32B. Specifically on Qwen3-VL-8B ($n{=}842$), $\rho = -0.733$ and $r = -0.573$ (both $p < 10^{-73}$), with mean \metric of $-0.678$ (95\% CI $[-0.738, -0.619]$, $p < 10^{-4}$) on correct predictions vs.\ $+0.046$ (95\% CI $[+0.021, +0.067]$, $p < 10^{-3}$) on incorrect ones, showing that these models lean on their language prior when they are wrong.
On POPE ($n{=}9{,}000$) and VQAv2 yes/no ($n{=}1{,}500$), \metric still predicts Qwen3-VL-8B errors above chance (AUROC $0.631$ and $0.640$). Note that we do expect this weaker separation since these datasets contain more irrelevant error-inducing factors other than language prior.

The significant correlation with performance and discriminative power of \metric indicates that the language prior bias, measured by \metric, is one of the meaningful root causes behind VLM failures.

\paragraph{Model-wise observations} \cref{tab:main_results} reports accuracy and \metric for each model. For every model the mean \metric is lower on correct predictions than on incorrect ones, with non-overlapping 95\% CIs. The correlation also holds within each model individually ($\rho$ from $-0.17$ to $-0.73$, all with $p < 10^{-6}$). Mean \metric is not directly comparable across models: the value function (\cref{eq:value_function}) is a KL difference whose scale depends on each model's own $p_{\text{LM}}$, and models differ in how many of their errors stem from the language prior. We therefore do not rank models by mean \metric and read differences across models only qualitatively. Within Qwen3-VL, the 8B and 32B models have lower \metric than the 2B and 4B ones ($-0.355$ and $-0.371$ vs. $-0.197$ and $-0.172$), indicating that model scale alleviates language prior bias within the Qwen3-VL series. At a similar size, Qwen3-VL-8B has much lower \metric than Qwen2-VL-7B and Qwen2.5-VL-7B ($-0.028$ and $-0.019$), suggesting that the training recipe can shift language-prior reliance independently of scale.

\begin{table}[t]
  \centering
  \small
  \caption{\textbf{Per-model accuracy and Language Prior Bias} on VLind-Bench-LP. Corr.\ and Incorr.\ give the mean \metric on correctly and incorrectly answered samples. Rows are ordered manually as we do not intend to compare or rank across models (\cref{sec:correlation_accuracy}).}
  \label{tab:main_results}
  \setlength{\tabcolsep}{3pt}
  \begin{tabular}{lcccc}
    \toprule
    \textbf{Model} & \textbf{Acc(\%)} $\uparrow$ & \textbf{\metric} $\downarrow$ & \textbf{Corr.} & \textbf{Incorr.} \\
    \midrule
    Qwen2-VL-7B    & 53.09          & $-0.028$ & $-0.074$ & $+0.025$ \\
    Qwen2.5-VL-7B  & 58.19          & $-0.019$ & $-0.062$ & $+0.041$ \\
    Qwen3-VL-2B    & 53.92          & $-0.197$ & $-0.314$ & $-0.060$ \\
    Qwen3-VL-4B    & 52.73          & $-0.172$ & $-0.389$ & $+0.071$ \\
    Qwen3-VL-8B    & 55.34          & $-0.355$ & $-0.678$ & $+0.046$ \\
    Qwen3-VL-32B   & 68.65          & $-0.371$ & $-0.562$ & $+0.046$ \\
    InternVL3-8B   & 59.62          & $-0.164$ & $-0.285$ & $+0.014$ \\
    Tarsier-7B     & 68.65          & $-0.060$ & $-0.070$ & $-0.039$ \\
    Molmo2-8B      & 71.02          & $-0.069$ & $-0.108$ & $+0.025$ \\
    \bottomrule
  \end{tabular}
\end{table}

\begin{table}[t]
  \centering
  \footnotesize
  \caption{\textbf{Comparison between \method and KernelSHAP on POPE and VLind-Bench-LP.} Wall-clock time is averaged across 5 POPE instances on one NVIDIA L40S. \#FWD is the total number of VLM calls needed. $r_{\text{\metric}}$ is the Pearson correlation of each method's \metric score and the KernelSHAP reference (computed with 1{,}500 coalition samples) over 64 POPE test samples and 84 VLind-Bench-LP validation samples.}
  \label{tab:efficiency}
  \begin{tabular}{lcccc}
    \toprule
    & & & \multicolumn{2}{c}{$\mathbf{r_{\text{\metric}}}$ $\uparrow$} \\
    \cmidrule(lr){4-5}
    \textbf{Method} & \textbf{Time (s)} & \textbf{\#FWD} & \textbf{POPE} & \textbf{VLind-LP} \\
    \midrule
    KernelSHAP-1,500 & 23.37 & 1{,}500 & 1.00 (ref.)     & 1.00 (ref.) \\
    KernelSHAP-850   & 13.42 &   850   & \textbf{0.9988} & \textbf{0.9990} \\
    KernelSHAP-200   & 3.30  &   200   & 0.9940          & 0.9955 \\
    \midrule
    \textbf{\method} & \textbf{0.13} & \textbf{2} & 0.9887 & 0.9914 \\
    \bottomrule
  \end{tabular}
  \vspace{-3mm}
\end{table}

\begin{table}[t]
  \centering
  \footnotesize
  \setlength{\tabcolsep}{1.5pt}
  \caption{\textbf{\metric as a fine-tuning objective} on VLind-Bench-LP. \emph{Base}: no fine-tuning; \emph{Ctrl}: control arm ($\lambda{=}0$); \emph{+\metric}: $\lambda{=}10$. Means over three seeds. Parentheses show the mean change of \emph{+\metric} relative to the same-seed \emph{Ctrl}; standard deviations across seeds and ablation on $\lambda$ are in \cref{app:finetune}.}
  \label{tab:finetune}
  \begin{tabular}{lcccccc}
    \toprule
    & \multicolumn{3}{c}{\textbf{Acc (\%)} $\uparrow$} & \multicolumn{3}{c}{\textbf{\metric} $\downarrow$} \\
    \cmidrule(lr){2-4} \cmidrule(lr){5-7}
    \textbf{Model} & Base & Ctrl & +\metric & Base & Ctrl & +\metric \\
    \midrule
    Qwen2-VL-7B                 & 59.7 & 60.9 & 68.1\,{\scriptsize($+7.2$)} & $-0.020$ & $-0.028$ & $-0.554$ \\
    Qwen2.5-VL-7B               & 63.2 & 62.6 & 62.2\,{\scriptsize($-0.4$)} & $-0.019$ & $-0.031$ & $-0.518$ \\
    Qwen3-VL-2B                 & 60.7 & 65.6 & 73.6\,{\scriptsize($+7.9$)} & $-0.147$ & $-0.076$ & $-0.798$ \\
    Qwen3-VL-4B                 & 59.9 & 60.6 & 66.6\,{\scriptsize($+6.0$)} & $-0.178$ & $-0.083$ & $-0.730$ \\
    Qwen3-VL-8B                 & 60.5 & 66.1 & 72.4\,{\scriptsize($+6.3$)} & $-0.341$ & $-0.131$ & $-1.044$ \\
    Tarsier-7B                  & 68.1 & 66.7 & 69.8\,{\scriptsize($+3.1$)} & $-0.063$ & $-0.026$ & $-0.165$ \\
    InternVL3-8B                & 60.6 & 62.3 & 65.7\,{\scriptsize($+3.4$)} & $-0.146$ & $-0.147$ & $-0.839$ \\
    Molmo2-8B                   & 72.2 & 71.3 & 73.5\,{\scriptsize($+2.2$)} & $-0.068$ & $-0.081$ & $-0.714$ \\
    \bottomrule
  \end{tabular}
  \vspace{-3mm}
\end{table}

\begin{figure*}[t]
    \centering
    \includegraphics[width=\linewidth]{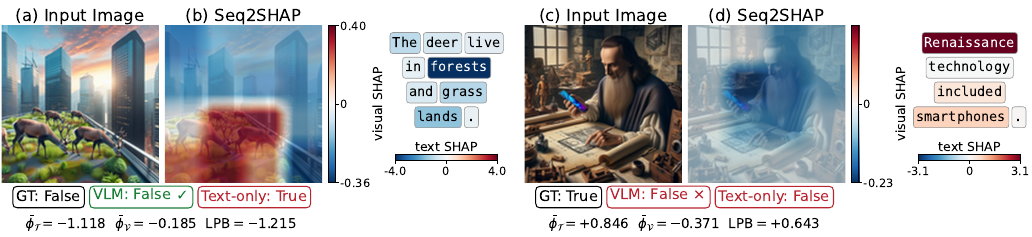}
    \vspace{-3mm}
    \caption{\textbf{Token-level \method attribution on two VLind-Bench-LP samples from Qwen3-VL-8B.}
        Token shading shows per-token text SHAP and the image overlay shows per-region visual SHAP. Red pushes the prediction toward the text-only prediction $p_{\text{LM}}$, and blue pushes it away. In both samples the text-only answer follows common sense and is wrong, so pulling away from it is the desired behavior. The model is correct on the left sample ($\text{\metric} = -0.61$) and wrong on the right ($\text{\metric} = +0.64$), so \metric separates the vision-grounded answer from the failure that follows the text prior. Discussion details in~\cref{sec:heatmaps}. More qualitative examples in~\cref{app:heatmaps}.}
    \label{fig:shap_heatmaps}
    \vspace{-3mm}
\end{figure*}

\subsection{Efficient Proxy to KernelSHAP}
\label{sec:vs_kernelshap}
\method replaces the thousands of coalition evaluations of KernelSHAP with only three passes: one standard VLM inference, which produces the prediction $p_{\mathcal{N}}$ and supplies the hidden states $\{\mathbf{h}_j\}_j$ for the explainer, one text-only pass for $p_{\text{LM}}$, and one forward pass of the explainer. \Cref{tab:efficiency} compares \method with KernelSHAP at three budgets on Qwen3-VL-8B. KernelSHAP-1{,}500 (KernelSHAP estimated with 1,500 coalition samples), is chosen as the baseline reference for its empirically stable Shapley estimates. \method takes \textbf{0.13}\,s per instance, a $\mathbf{187\times}$ speedup over KernelSHAP-1{,}500 (23.37\,s) and $26\times$ over KernelSHAP-200 (3.30\,s). Yet its predicted \metric scores correlate with the reference at $r_{\text{\metric}} = \textbf{0.9887}$ on POPE and \textbf{0.9914} on VLind-Bench-LP, within 0.6\% of KernelSHAP-200 on both datasets, reasonably approximating the baseline \metric.

\subsection{\metric Guides Model Improvement}
\label{sec:actionable}

Building on the correlation findings in~\cref{sec:correlation_accuracy}, we show that \metric is actionable, \ie, whether lowering it makes a model more accurate. The value function (\cref{eq:value_function}) needs no ground-truth label and the vision term of \cref{eq:lpb} is differentiable, so \metric can be added directly to the VLM training loss.

\paragraph{Setup} We LoRA-fine-tune all of the VLMs (except Qwen3-VL-32B) on the yes/no subset of VQAv2~\cite{goyal2017making} with loss $\mathcal{L} = \mathcal{L}_{\text{CE}} + \lambda\,\mathcal{L}_\text{\metric}$ ($\mathcal{L}_{\text{CE}}$ for cross-entropy loss; $\lambda{=}10$ with ablation in \cref{app:finetune}), and evaluate on VLind-Bench-LP, so training and evaluation samples are from different datasets. A control arm with the same recipe, seed, and $\lambda{=}0$ separates the effect of the \metric term from ordinary fine-tuning gains. To keep the loss exact and cheap to compute, training uses the modality-level \metric (\cref{sec:lpb_metric}): the two-player Shapley values $\psi_{\mathcal{T}}, \psi_{\mathcal{V}}$ come from four forward passes per sample (all players, text only, vision only, none) and are divided by $N_{\mathcal{T}}$ and $N_{\mathcal{V}}$. 
Training details are in \cref{app:finetune}.

\paragraph{Results} Relative to the same-seed control, the \metric term raises accuracy by $4.5$ points on average over three seeds, with gains on seven of eight models, and lowers \metric on all eight (\cref{tab:finetune}). Ordinary fine-tuning alone does not achieve this: the control arm improves on the base model by only $1.4$ points on average. Gains are largest on the Qwen2-VL and Qwen3-VL models ($+6.0$ to $+7.9$) and smallest on Tarsier-7B and Molmo2-8B ($+3.1$ and $+2.2$), the two models with the highest control accuracy. Although training and evaluation data differ, minimizing \metric transfers to accuracy gains on the VLind-Bench-LP benchmark for seven of eight models. Cross-seed variance and ablation on $\lambda$ are shown in~\cref{app:finetune}.

\paragraph{Prompt refinement} \metric can also be optimized at deployment time without changing model weights. For each sample, a genetic search over instruction prefixes keeps the candidate with the lowest \metric. We use gpt-5.4-mini~\cite{openai2026gpt54mini} to proposes the prefixes, and it sees only past prefixes and their \metric scores, never the image, question, or label, so any gain must come through \metric. Candidates are scored with a trained \method explainer, which keeps the \metric evaluations per sample affordable. On the $374$ VLind-Bench-LP samples that Qwen3-VL-8B gets wrong, the search corrects $113$ ($30.2\%$), vs.\ $7$ ($1.9\%$) for the seed prefix without search. The flips go both ways ($73$ True$\to$False, $40$ False$\to$True), so the prefixes do not simply bias the answer; the best ones are short, generic instructions to look at the image. More details are in~\cref{app:prompt_refinement}.

\subsection{Qualitative \method Heatmaps}
\label{sec:heatmaps}
\cref{fig:shap_heatmaps} shows two VLind-Bench-LP samples, each pairing a counter-commonsense image with a statement the model must judge True or False. In VLind-Bench-LP, each image depicts a counterfactual scene that contradicts common sense, and the model must judge a statement True or False from the image alone. Answering correctly therefore requires vision to override the language prior. In both, the text-only prediction follows common sense and is wrong, so a correct answer requires vision to override the language prior. Qwen3-VL-8B overrides the prior on the left and follows it on the right. Positive attributions pull the output toward $p_{\text{LM}}$ and negative ones pull it away. On the left, the attributions pull away from the prior and the model answers correctly ($\text{\metric}=-0.61$). On the right, they fail to pull away, and the model wrongly agrees with the text-only prediction ($\text{\metric}=+0.64$).

\section{Discussion}
\label{sec:discussion}

\paragraph{When \metric is misleading}
\metric is meaningful only when the correct answer requires visual input that cannot be inferred from the question alone, which is precisely the design principle of VLind-Bench-LP. On questions that genuinely call for textual reasoning (e.g., ``What color is associated with stop signs?''), high \metric reflects appropriate behavior rather than bias. Even on prior-conflict data, language prior bias is one failure mode among several (\eg, perceptual or compositional errors), so \metric is not expected to correlate perfectly with accuracy. We recommend interpreting \metric in the context of the specific tasks instead of as a blanket evaluation of VLMs.

\paragraph{Limitations}
Our empirical study covers two single-image yes/no benchmarks, POPE and VLind-Bench-LP, chosen because they most cleanly isolate language-prior bias. Such controlled prior-conflict benchmarks are scarce: VQA-CP~\cite{agrawal2018vqacp} shifts answer distributions between its train and test splits but does not pair individual questions with prior-contradicting images, while Winoground~\cite{thrush2022winoground} and MMVP~\cite{tong2024mmvp} target compositionality and perception rather than prior conflict specifically. We do not yet report numbers on multi-way MCQ, video, or open-ended generation. These are not gaps in the method: the Shapley game already handles $C$-way option spaces (yes/no is $C{=}2$), and the explainer, value function, and \metric aggregation carry over to video with temporal tokens as players. Since the game requires no ground-truth answer, they also extend to open-ended generation (\cref{app:extension}). Empirical validation of these extensions is left to future work. We also note that since \metric is designed to detect prior-driven errors, we do not expect it to explain all failures: a prediction can also be wrong for reasons unrelated to the language prior, such as perception errors or uncertainty. Finally, we verify symmetry and dummy only approximately (\cref{app:axioms}) and do not test linearity, which would require training and combining separate explainers.

\paragraph{Broader implications}
Since \metric requires no ground-truth answer, it applies wherever labels are unavailable. Per sample, it can flag predictions likely driven by the language prior, letting a deployed system abstain or defer them to human review. Per dataset, high \metric on correctly answered questions marks items answerable without the image, offering a way to audit benchmarks for text-only leakage~\cite{chen2024mmstar}. More broadly, with language prior bias now quantifiable, \metric makes it a target that future VLMs can be designed against, \eg, as a training objective (\cref{sec:actionable}).

\section{Conclusion}
\label{sec:conclusion}

Language prior bias has long been recognized as a fundamental weakness of VLMs, yet prior work has largely stopped at detection: flagging the symptom without measuring the cause. We close this gap with \metric, a Shapley-value-based metric that quantifies, at token-level granularity, how much each prediction is driven by visual evidence versus language priors, turning a qualitative concern into an interpretable, actionable number.

Realizing this metric at the scale of modern VLMs required overcoming a hard computational barrier: exact Shapley estimation costs thousands of forward passes per instance. We resolve it with \method, the first amortized Shapley method for autoregressive multimodal models, which collapses this cost to two VLM inferences plus a single forward pass through a lightweight explainer.

Through our experiments, \metric provides quantitative evidence for the hypothesis that language prior bias impairs prediction: across nine VLMs from the Qwen-VL, Molmo2, Tarsier, and InternVL3 families on VLind-Bench-LP, predictions that rely more on language priors are systematically less accurate with pooled Spearman $\rho = -0.568$ between \metric and accuracy, making \metric a label-free indicator of when a prediction is likely to be wrong (AUROC $= 0.835$). Crucially, \metric is not just descriptive but \emph{actionable}: as a fine-tuning regularizer, \metric reduces bias on all eight models and improves accuracy by $+4.5$ points on average, and \metric is also shown as effective prompt refinement guidance, opening the door to more future \metric-aware model design.

As VLMs are increasingly deployed in settings where people act on their answers, being correct is not enough: a trustworthy VLM should answer from what it sees, not from what its language prior expects. We see this work as a step from \emph{detecting} such failures to \emph{quantifying and intervening on} them. Since \metric requires no ground-truth label, the score reaches beyond the benchmarks studied here: it can flag prior-driven answers at deployment, audit benchmarks for questions answerable from text alone, and guide the design of future models. We will release \method and \metric as open-source tools to facilitate building VLMs that are not only accurate, but trustworthy.

\section*{Acknowledgments}
We used Claude (Anthropic) to assist with wording, code, and literature search; the authors verified all content and take full responsibility for it.

{
    \small
    \bibliographystyle{ieeenat_fullname}
    \bibliography{references}

\begin{thebibliography}{57}
\providecommand{\natexlab}[1]{#1}
\providecommand{\url}[1]{\texttt{#1}}
\expandafter\ifx\csname urlstyle\endcsname\relax
  \providecommand{\doi}[1]{doi: #1}\else
  \providecommand{\doi}{doi: \begingroup \urlstyle{rm}\Url}\fi

\bibitem[Agrawal et~al.(2018)Agrawal, Batra, Parikh, and Kembhavi]{agrawal2018vqacp}
Aishwarya Agrawal, Dhruv Batra, Devi Parikh, and Aniruddha Kembhavi.
\newblock Don't just assume; look and answer: Overcoming priors for visual question answering.
\newblock In \emph{Proceedings of the IEEE conference on computer vision and pattern recognition}, pages 4971--4980, 2018.

\bibitem[Ainslie et~al.(2023)Ainslie, Lee-Thorp, De~Jong, Zemlyanskiy, Lebr{\'o}n, and Sanghai]{ainslie2023gqa}
Joshua Ainslie, James Lee-Thorp, Michiel De~Jong, Yury Zemlyanskiy, Federico Lebr{\'o}n, and Sumit Sanghai.
\newblock Gqa: Training generalized multi-query transformer models from multi-head checkpoints.
\newblock In \emph{Proceedings of the 2023 conference on empirical methods in natural language processing}, pages 4895--4901, 2023.

\bibitem[Alayrac et~al.(2022)Alayrac, Donahue, Luc, Miech, Barr, Hasson, Lenc, Mensch, Millican, Reynolds, et~al.]{alayrac2022flamingo}
Jean-Baptiste Alayrac, Jeff Donahue, Pauline Luc, Antoine Miech, Iain Barr, Yana Hasson, Karel Lenc, Arthur Mensch, Katherine Millican, Malcolm Reynolds, et~al.
\newblock Flamingo: a visual language model for few-shot learning.
\newblock \emph{Advances in neural information processing systems}, 35:\penalty0 23716--23736, 2022.

\bibitem[Bai et~al.(2025{\natexlab{a}})Bai, Cai, Chen, Chen, Chen, Cheng, Deng, Ding, Gao, Ge, et~al.]{bai2025qwen3vl}
Shuai Bai, Yuxuan Cai, Ruizhe Chen, Keqin Chen, Xionghui Chen, Zesen Cheng, Lianghao Deng, Wei Ding, Chang Gao, Chunjiang Ge, et~al.
\newblock {Qwen3-VL} technical report, 2025{\natexlab{a}}.

\bibitem[Bai et~al.(2025{\natexlab{b}})Bai, Chen, Liu, Wang, Ge, Song, Dang, Wang, Wang, Tang, et~al.]{bai2025qwen25vl}
Shuai Bai, Keqin Chen, Xuejing Liu, Jialin Wang, Wenbin Ge, Sibo Song, Kai Dang, Peng Wang, Shijie Wang, Jun Tang, et~al.
\newblock {Qwen2.5-VL} technical report, 2025{\natexlab{b}}.

\bibitem[Basu et~al.(2024)Basu, Grayson, Morrison, Nushi, Feizi, and Massiceti]{basu2024understanding}
Samyadeep Basu, Martin Grayson, Cecily Morrison, Besmira Nushi, Soheil Feizi, and Daniela Massiceti.
\newblock Understanding information storage and transfer in multi-modal large language models.
\newblock In \emph{Proceedings of the 38th International Conference on Neural Information Processing Systems}, pages 7400--7426, 2024.

\bibitem[Cadene et~al.(2019)Cadene, Dancette, Cord, Parikh, et~al.]{cadene2019rubi}
Remi Cadene, Corentin Dancette, Matthieu Cord, Devi Parikh, et~al.
\newblock {RUBi}: Reducing unimodal biases for visual question answering.
\newblock In \emph{Advances in Neural Information Processing Systems (NeurIPS)}, 2019.

\bibitem[Cappellazzo et~al.(2026)Cappellazzo, Petridis, and Pantic]{cappellazzo2026drshapav}
Umberto Cappellazzo, Stavros Petridis, and Maja Pantic.
\newblock Dr. shap-av: Decoding relative modality contributions via shapley attribution in audio-visual speech recognition, 2026.

\bibitem[Chakraborty et~al.(2026)Chakraborty, Wang, Zhang, Bai, Yu, Kampffmeyer, Lee, De~la Torre, and Jenssen]{chakraborty2026notanother}
Rwiddhi Chakraborty, Yinong~Oliver Wang, Cheng Zhang, Fan Bai, Zhuoran Yu, Michael Kampffmeyer, Yong~Jae Lee, Fernando De~la Torre, and Robert Jenssen.
\newblock Not another text benchmark: Putting the ``visual'' back in visual question answering for large video models.
\newblock In \emph{Advances in Neural Information Processing Systems (NeurIPS)}, 2026.

\bibitem[Chen et~al.(2023)Chen, Lou, Zhang, Huang, and Zhang]{chen2023harsanyinet}
Lu Chen, Siyu Lou, Keyan Zhang, Jin Huang, and Quanshi Zhang.
\newblock Harsanyinet: Computing accurate shapley values in a single forward propagation.
\newblock \emph{arXiv preprint arXiv:2304.01811}, 2023.

\bibitem[Chen et~al.(2024)Chen, Li, Dong, Zhang, Zang, Chen, Duan, Wang, Qiao, Lin, et~al.]{chen2024mmstar}
Lin Chen, Jinsong Li, Xiaoyi Dong, Pan Zhang, Yuhang Zang, Zehui Chen, Haodong Duan, Jiaqi Wang, Yu Qiao, Dahua Lin, et~al.
\newblock Are we on the right way for evaluating large vision-language models?
\newblock In \emph{Advances in Neural Information Processing Systems (NeurIPS)}, 2024.

\bibitem[Clark et~al.(2026)]{clark2026molmo2}
Christopher Clark et~al.
\newblock {Molmo2}: Open weights and data for vision-language models with video understanding and grounding.
\newblock arXiv preprint, 2026.

\bibitem[Covert et~al.(2024)Covert, Kim, Lee, Zou, and Hashimoto]{covert2024stochastic}
Ian Covert, Chanwoo Kim, Su-In Lee, James Zou, and Tatsunori Hashimoto.
\newblock Stochastic amortization: A unified approach to accelerate feature and data attribution.
\newblock In \emph{Advances in Neural Information Processing Systems}, 2024.

\bibitem[Covert et~al.(2020)Covert, Lundberg, and Lee]{covert2020sage}
Ian~C Covert, Scott Lundberg, and Su-In Lee.
\newblock Understanding global feature contributions with additive importance measures.
\newblock In \emph{Proceedings of the 34th International Conference on Neural Information Processing Systems}, pages 17212--17223, 2020.

\bibitem[Dai et~al.(2023)Dai, Li, Li, Tiong, Zhao, Wang, Li, Fung, and Hoi]{dai2023instructblip}
Wenliang Dai, Junnan Li, Dongxu Li, Anthony Tiong, Junqi Zhao, Weisheng Wang, Boyang Li, Pascale Fung, and Steven Hoi.
\newblock {InstructBLIP}: Towards general-purpose vision-language models with instruction tuning.
\newblock In \emph{Advances in Neural Information Processing Systems (NeurIPS)}, 2023.

\bibitem[Enouen et~al.(2024)Enouen, Nakhost, Ebrahimi, Arik, Liu, and Pfister]{enouen2024textgenshap}
James Enouen, Hootan Nakhost, Sayna Ebrahimi, Sercan Arik, Yan Liu, and Tomas Pfister.
\newblock Textgenshap: Scalable post-hoc explanations in text generation with long documents.
\newblock In \emph{Findings of the Association for Computational Linguistics: ACL 2024}, pages 13984--14011, 2024.

\bibitem[Frye et~al.(2020)Frye, Rowat, and Feige]{frye2020asymmetric}
Christopher Frye, Colin Rowat, and Ilya Feige.
\newblock Asymmetric shapley values: incorporating causal knowledge into model-agnostic explainability.
\newblock In \emph{Advances in Neural Information Processing Systems (NeurIPS)}, 2020.

\bibitem[Goyal et~al.(2017)Goyal, Khot, Summers-Stay, Batra, and Parikh]{goyal2017making}
Yash Goyal, Tejas Khot, Douglas Summers-Stay, Dhruv Batra, and Devi Parikh.
\newblock Making the v in vqa matter: Elevating the role of image understanding in visual question answering.
\newblock In \emph{Proceedings of the IEEE conference on computer vision and pattern recognition}, pages 6904--6913, 2017.

\bibitem[Guan et~al.(2024)Guan, Liu, Wu, Xian, Li, Liu, Wang, Chen, Huang, Yacoob, et~al.]{guan2024hallusionbench}
Tianrui Guan, Fuxiao Liu, Xiyang Wu, Ruiqi Xian, Zongxia Li, Xiaoyu Liu, Xijun Wang, Lichang Chen, Furong Huang, Yaser Yacoob, et~al.
\newblock Hallusionbench: an advanced diagnostic suite for entangled language hallucination and visual illusion in large vision-language models.
\newblock In \emph{Proceedings of the IEEE/CVF conference on computer vision and pattern recognition}, pages 14375--14385, 2024.

\bibitem[Jethani et~al.(2022)Jethani, Sudarshan, Covert, Lee, and Ranganath]{jethani2022fastshap}
Neil Jethani, Mukund Sudarshan, Ian~Connick Covert, Su-In Lee, and Rajesh Ranganath.
\newblock Fastshap: Real-time shapley value estimation.
\newblock In \emph{International conference on learning representations}, 2022.

\bibitem[Jethani et~al.(2023)Jethani, Saporta, and Ranganath]{jethani2023label}
Neil Jethani, Adriel Saporta, and Rajesh Ranganath.
\newblock Don’t be fooled: label leakage in explanation methods and the importance of their quantitative evaluation.
\newblock In \emph{International Conference on Artificial Intelligence and Statistics}, pages 8925--8953. PMLR, 2023.

\bibitem[Kervadec et~al.(2021)Kervadec, Antipov, Baccouche, and Wolf]{kervadec2021roses}
Corentin Kervadec, Grigory Antipov, Moez Baccouche, and Christian Wolf.
\newblock Roses are red, violets are blue... but should vqa expect them to?
\newblock In \emph{Proceedings of the IEEE/CVF Conference on Computer Vision and Pattern Recognition}, pages 2776--2785, 2021.

\bibitem[Krojer et~al.(2025)Krojer, Komeili, Ross, Garrido, Sinha, Ballas, and Assran]{krojer2025mvp}
Benno Krojer, Mojtaba Komeili, Candace Ross, Quentin Garrido, Koustuv Sinha, Nicolas Ballas, and Mahmoud Assran.
\newblock A shortcut-aware video-qa benchmark for physical understanding via minimal video pairs.
\newblock \emph{arXiv preprint arXiv:2506.09987}, 2025.

\bibitem[Lee et~al.(2025)Lee, Kim, Yoon, Kim, Lee, Koh, and Jung]{lee2024vlindbench}
Kang-il Lee, Minbeom Kim, Seunghyun Yoon, Minsung Kim, Dongryeol Lee, Hyukhun Koh, and Kyomin Jung.
\newblock Vlind-bench: Measuring language priors in large vision-language models.
\newblock In \emph{Findings of the Association for Computational Linguistics: NAACL 2025}, pages 4129--4144, 2025.

\bibitem[Leng et~al.(2024)Leng, Zhang, Chen, Li, Lu, Miao, and Bing]{leng2024vcd}
Sicong Leng, Hang Zhang, Guanzheng Chen, Xin Li, Shijian Lu, Chunyan Miao, and Lidong Bing.
\newblock Mitigating object hallucinations in large vision-language models through visual contrastive decoding.
\newblock In \emph{Proceedings of the IEEE/CVF Conference on Computer Vision and Pattern Recognition}, pages 13872--13882, 2024.

\bibitem[Li et~al.(2023)Li, Du, Zhou, Wang, Zhao, and Wen]{li2023pope}
Yifan Li, Yifan Du, Kun Zhou, Jinpeng Wang, Xin Zhao, and Ji-Rong Wen.
\newblock Evaluating object hallucination in large vision-language models.
\newblock In \emph{Proceedings of the 2023 conference on empirical methods in natural language processing}, pages 292--305, 2023.

\bibitem[Liang et~al.(2021)Liang, Hu, and Zhu]{liang2021lpf}
Zujie Liang, Haifeng Hu, and Jiaying Zhu.
\newblock Lpf: A language-prior feedback objective function for de-biased visual question answering.
\newblock In \emph{Proceedings of the 44th international ACM SIGIR conference on research and development in information retrieval}, pages 1955--1959, 2021.

\bibitem[Liu et~al.(2023)Liu, Li, Wu, and Lee]{liu2024llava}
Haotian Liu, Chunyuan Li, Qingyang Wu, and Yong~Jae Lee.
\newblock Visual instruction tuning.
\newblock In \emph{Advances in Neural Information Processing Systems (NeurIPS)}, 2023.

\bibitem[Long et~al.(2025)Long, Oh, Park, and Li]{long2025understanding}
Lin Long, Changdae Oh, Seongheon Park, and Sharon Li.
\newblock Understanding language prior of lvlms by contrasting chain-of-embedding.
\newblock \emph{arXiv preprint arXiv:2509.23050}, 2025.

\bibitem[Lundberg and Lee(2017)]{lundberg2017unified}
Scott~M Lundberg and Su-In Lee.
\newblock A unified approach to interpreting model predictions.
\newblock In \emph{Advances in neural information processing systems}, 2017.

\bibitem[Luo et~al.(2025)Luo, Cao, Lee, Johnson, and Lee]{luo2025probing}
Tiange Luo, Ang Cao, Gunhee Lee, Justin Johnson, and Honglak Lee.
\newblock Probing visual language priors in vlms.
\newblock In \emph{International Conference on Machine Learning}, pages 41120--41156. PMLR, 2025.

\bibitem[{OpenAI}(2026)]{openai2026gpt54mini}
{OpenAI}.
\newblock Introducing {GPT-5.4} mini and nano.
\newblock \url{https://openai.com/index/introducing-gpt-5-4-mini-and-nano/}, 2026.

\bibitem[Parcalabescu and Frank(2023)]{parcalabescu2023mmshap}
Letitia Parcalabescu and Anette Frank.
\newblock Mm-shap: A performance-agnostic metric for measuring multimodal contributions in vision and language models \& tasks.
\newblock In \emph{Proceedings of the 61st Annual Meeting of the Association for Computational Linguistics (Volume 1: Long Papers)}, pages 4032--4059, 2023.

\bibitem[Parcalabescu and Frank(2025)]{parcalabescu2025decoders}
Letitia Parcalabescu and Anette Frank.
\newblock Do vision \& language decoders use images and text equally? how self-consistent are their explanations?
\newblock In \emph{International Conference on Learning Representations (ICLR)}, 2025.

\bibitem[Park et~al.(2025)Park, Jang, Alasaly, Mopidevi, Zolensky, Eaton, Lee, and Johnson]{park2024modality}
Jean Park, Kuk~Jin Jang, Basam Alasaly, Sriharsha Mopidevi, Andrew Zolensky, Eric Eaton, Insup Lee, and Kevin Johnson.
\newblock Assessing modality bias in video question answering benchmarks with multimodal large language models.
\newblock In \emph{Proceedings of the AAAI Conference on Artificial Intelligence}, pages 19821--19829, 2025.

\bibitem[Pollard and Wray(2025)]{pollard2025videoworththousandwords}
Sam Pollard and Michael Wray.
\newblock A video is not worth a thousand words, 2025.

\bibitem[Radford et~al.(2021)Radford, Kim, Hallacy, Ramesh, Goh, Agarwal, Sastry, Askell, Mishkin, Clark, et~al.]{radford2021clip}
Alec Radford, Jong~Wook Kim, Chris Hallacy, Aditya Ramesh, Gabriel Goh, Sandhini Agarwal, Girish Sastry, Amanda Askell, Pamela Mishkin, Jack Clark, et~al.
\newblock Learning transferable visual models from natural language supervision.
\newblock In \emph{International conference on machine learning}, pages 8748--8763. PmLR, 2021.

\bibitem[Rohrbach et~al.(2018)Rohrbach, Hendricks, Burns, Darrell, and Saenko]{rohrbach2018chair}
Anna Rohrbach, Lisa~Anne Hendricks, Kaylee Burns, Trevor Darrell, and Kate Saenko.
\newblock Object hallucination in image captioning.
\newblock In \emph{Proceedings of the 2018 Conference on Empirical Methods in Natural Language Processing}, pages 4035--4045, 2018.

\bibitem[Shapley(1953)]{shapley1953value}
Lloyd~S. Shapley.
\newblock A value for n-person games.
\newblock In \emph{Contributions to the Theory of Games II}, pages 307--317. Princeton University Press, 1953.

\bibitem[Shazeer(2020)]{shazeer2020glu}
Noam Shazeer.
\newblock Glu variants improve transformer.
\newblock \emph{arXiv preprint arXiv:2002.05202}, 2020.

\bibitem[{\v{S}}trumbelj and Kononenko(2014)]{strumbelj2014explaining}
Erik {\v{S}}trumbelj and Igor Kononenko.
\newblock Explaining prediction models and individual predictions with feature contributions.
\newblock \emph{Knowledge and information systems}, 41\penalty0 (3):\penalty0 647--665, 2014.

\bibitem[Thrush et~al.(2022)Thrush, Jiang, Bartolo, Singh, Williams, Kiela, and Ross]{thrush2022winoground}
Tristan Thrush, Ryan Jiang, Max Bartolo, Amanpreet Singh, Adina Williams, Douwe Kiela, and Candace Ross.
\newblock Winoground: Probing vision and language models for visio-linguistic compositionality.
\newblock In \emph{Proceedings of the IEEE/CVF Conference on Computer Vision and Pattern Recognition}, pages 5238--5248, 2022.

\bibitem[Tong et~al.(2024)Tong, Liu, Zhai, Ma, LeCun, and Xie]{tong2024mmvp}
Shengbang Tong, Zhuang Liu, Yuexiang Zhai, Yi Ma, Yann LeCun, and Saining Xie.
\newblock Eyes wide shut? exploring the visual shortcomings of multimodal llms.
\newblock In \emph{Proceedings of the IEEE/CVF conference on computer vision and pattern recognition}, pages 9568--9578, 2024.

\bibitem[Wang et~al.(2024{\natexlab{a}})Wang, Yuan, Zhang, and Sun]{wang2024tarsier}
Jiawei Wang, Liping Yuan, Yuchen Zhang, and Haomiao Sun.
\newblock Tarsier: Recipes for training and evaluating large video description models, 2024{\natexlab{a}}.

\bibitem[Wang et~al.(2024{\natexlab{b}})Wang, Bai, Tan, Wang, Fan, Bai, Chen, Liu, Wang, Ge, et~al.]{wang2024qwen2}
Peng Wang, Shuai Bai, Sinan Tan, Shijie Wang, Zhihao Fan, Jinze Bai, Keqin Chen, Xuejing Liu, Jialin Wang, Wenbin Ge, et~al.
\newblock Qwen2-vl: Enhancing vision-language model's perception of the world at any resolution.
\newblock \emph{arXiv preprint arXiv:2409.12191}, 2024{\natexlab{b}}.

\bibitem[Wang et~al.(2024{\natexlab{c}})Wang, Pan, Ding, and Biemann]{wang2024icd}
Xintong Wang, Jingheng Pan, Liang Ding, and Chris Biemann.
\newblock Mitigating hallucinations in large vision-language models with instruction contrastive decoding.
\newblock In \emph{Findings of the Association for Computational Linguistics: ACL 2024}, pages 15840--15853, 2024{\natexlab{c}}.

\bibitem[Wang and Wang(2025)]{wang2025multishap}
Zhanliang Wang and Kai Wang.
\newblock Multishap: A shapley-based framework for explaining cross-modal interactions in multimodal ai models, 2025.

\bibitem[Wang et~al.(2026)Wang, He, Li, Yang, Xiong, and Liu]{wang2026vfat}
Ziteng Wang, Yujie He, Guanliang Li, Siqi Yang, Jiaqi Xiong, and Songxiang Liu.
\newblock V-fat: Benchmarking visual fidelity against text-bias, 2026.

\bibitem[Wu et~al.(2025{\natexlab{a}})Wu, Tang, Zheng, and Jiang]{wu2025language}
Huyu Wu, Meng Tang, Xinhan Zheng, and Haiyun Jiang.
\newblock When language overrules: Revealing text dominance in multimodal large language models.
\newblock \emph{arXiv preprint arXiv:2508.10552}, 2025{\natexlab{a}}.

\bibitem[Wu et~al.(2025{\natexlab{b}})Wu, Sun, Wen, and Haber]{wu2025svco}
Shengguang Wu, Fan-Yun Sun, Kaiyue Wen, and Nick Haber.
\newblock Symmetrical visual contrastive optimization: Aligning vision-language models with minimal contrastive images.
\newblock In \emph{Proceedings of the 63rd Annual Meeting of the Association for Computational Linguistics (Volume 1: Long Papers)}, pages 30284--30297, 2025{\natexlab{b}}.

\bibitem[Xiao et~al.(2021)Xiao, Shang, Yao, and Chua]{xiao2021nextqa}
Junbin Xiao, Xindi Shang, Angela Yao, and Tat-Seng Chua.
\newblock Next-qa: Next phase of question-answering to explaining temporal actions.
\newblock In \emph{Proceedings of the IEEE/CVF conference on computer vision and pattern recognition}, pages 9777--9786, 2021.

\bibitem[Xiong et~al.(2020)Xiong, Yang, He, Zheng, Zheng, Xing, Zhang, Lan, Wang, and Liu]{xiong2020layer}
Ruibin Xiong, Yunchang Yang, Di He, Kai Zheng, Shuxin Zheng, Chen Xing, Huishuai Zhang, Yanyan Lan, Liwei Wang, and Tieyan Liu.
\newblock On layer normalization in the transformer architecture.
\newblock In \emph{International conference on machine learning}, pages 10524--10533. PMLR, 2020.

\bibitem[Yang et~al.(2023{\natexlab{a}})Yang, Yin, He, Chang, Ma, and Xiang]{yang2023amortized}
Chenghao Yang, Fan Yin, He He, Kai-Wei Chang, Xiaofei Ma, and Bing Xiang.
\newblock Efficient shapley values estimation by amortization for text classification.
\newblock In \emph{Proceedings of the 61st Annual Meeting of the Association for Computational Linguistics (Volume 1: Long Papers)}, pages 8666--8680, 2023{\natexlab{a}}.

\bibitem[Yang et~al.(2023{\natexlab{b}})Yang, Yin, He, Chang, Ma, and Xiang]{yang2023efficient}
Chenghao Yang, Fan Yin, He He, Kai-Wei Chang, Xiaofei Ma, and Bing Xiang.
\newblock Efficient {Shapley} values estimation by amortization for text classification.
\newblock In \emph{Proceedings of the 61st Annual Meeting of the Association for Computational Linguistics (Volume 1: Long Papers)}, pages 8666--8680, 2023{\natexlab{b}}.

\bibitem[Zhang and Sennrich(2019)]{zhang2019rmsnorm}
Biao Zhang and Rico Sennrich.
\newblock Root mean square layer normalization.
\newblock In \emph{Advances in neural information processing systems}, 2019.

\bibitem[Zhou et~al.(2024)Zhou, Cui, Yoon, Zhang, Deng, Finn, Bansal, and Yao]{zhou2024lure}
Yiyang Zhou, Chenhang Cui, Jaehong Yoon, Linjun Zhang, Zhun Deng, Chelsea Finn, Mohit Bansal, and Huaxiu Yao.
\newblock Analyzing and mitigating object hallucination in large vision-language models.
\newblock In \emph{The Twelfth International Conference on Learning Representations}, 2024.

\bibitem[Zhu et~al.(2025)Zhu, Wang, Chen, Liu, Ye, Gu, Tian, Duan, Su, Shao, et~al.]{zhu2025internvl3}
Jinguo Zhu, Weiyun Wang, Zhe Chen, Zhaoyang Liu, Shenglong Ye, Lixin Gu, Hao Tian, Yuchen Duan, Weijie Su, Jie Shao, et~al.
\newblock Internvl3: Exploring advanced training and test-time recipes for open-source multimodal models.
\newblock \emph{arXiv preprint arXiv:2504.10479}, 2025.

\end{thebibliography}
}

\clearpage
\maketitlesupplementary
\appendix
\section{Formulation Extensions}
\label[appendix]{app:extension}

\subsection{Multi-Token Open-Ended Evaluation}

The main paper scores the cooperative game on a single answer-position token, which matches the canonical evaluation protocol of POPE and VLind-Bench-LP. This section shows that the formulation, the coalition cache, and the \method explainer all extend naturally to multi-token open-ended outputs, and characterizes the additional cost.

\paragraph{Per-token cooperative game}
Let $y^\star = (y_1^\star, \ldots, y_T^\star)$ be the model's greedy decode on the full input, and let $y_{<t}^\star = (y_1^\star, \ldots, y_{t-1}^\star)$ denote its first $t-1$ tokens (empty for $t=1$). Let $p_S^{(t)}$ denote the model's next-token distribution at position $t$ given $\mathbf{x}_S$ followed by $y_{<t}^\star$ (with $p_\emptyset^{(t)}$ the null-coalition case), let $p_{\text{LM}}^{(t)}$ denote its counterpart on the text-only input followed by $y_{<t}^\star$, and let $g_t(\cdot) = \mathrm{KL}\left(p_{\text{LM}}^{(t)} \,\|\, \cdot\right)$. For each output position $t$ we define the per-token value function
\begin{equation}
\label{eq:multitoken_value}
v_t(S) \;=\; g_t\big(p_\emptyset^{(t)}\big) - g_t\big(p_S^{(t)}\big),
\end{equation}
analogous to \cref{eq:value_function}. Since open-ended outputs have no option set, all distributions are taken over the full vocabulary rather than renormalized over $C$ option tokens. The game stays label-free: $y_t^\star$ is never scored as a target and enters only as part of the prefix for later positions. The decoded prefix $y_{<t}^\star$ is fixed per sample (it depends on the full input, not on $S$), so each output position induces a separate cooperative game over the same player set $\mathcal{N}$ and the same coalition draws. The per-token Shapley vector $\boldsymbol{\phi}^{(t)} \in \mathbb{R}^{|\mathcal{N}|}$ produces an $|\mathcal{N}| \times T$ attribution matrix.

\paragraph{Compute cost}
The natural worry is that producing $T$ values multiplies cost by $T$. It does not, because all $T$ next-token distributions can be obtained from a single teacher-forced prefill on $[x_S, y_{<T}^\star]$. Concretely, for each coalition $S$:
\begin{enumerate}
    \item Apply the attention-mask coalition only to the input positions in $\mathcal{N}$ and keep the appended decoded tokens $y_{<T}^\star$ as always-on.
    \item Run a single forward pass. Read off $p_S^{(t)}$ from the logits at each output position $t$.
\end{enumerate}
Per-coalition cost grows from a length-$L$ prefill to a length-$(L{+}T)$ prefill, an overhead of $(L{+}T)/L$. Because $L$ is dominated by the visual stream while $T$ is dominated by the verbal answer (typically $T \in [5, 50]$), this overhead is at most a few percent. The text-only distributions $p_{\text{LM}}^{(t)}$ need one extra teacher-forced pass per sample, not per coalition.

\paragraph{Cache and explainer changes}
The coalition cache stores the value $v_t(S)$ at each output position rather than only at the answer position, so the full-vocabulary distributions need not be stored. Storage grows linearly in $T$, to $K \times T$ floats per sample, which is of the same order as the coalition mask itself ($K \times L$ bits) for typical answer lengths. The \method explainer head produces a $T$-vector per token rather than a scalar, so only the input projection and the head change and the encoder is unaffected. The divergence input $\kappa$ likewise becomes a per-position $\kappa_t = g_t\big(p_{\mathcal{N}}^{(t)}\big)$. The KernelSHAP targets $\boldsymbol{\phi}^*$ (\cref{eq:phi_star}) and the loss (\cref{eq:training_loss}) apply per output position.


\paragraph{Aggregation into \metric}
The \metric metric (\cref{sec:lpb_metric}) extends to multi-token outputs by computing $\bar{\phi}_{\mathcal{T}}^{(t)}$, $\bar{\phi}_{\mathcal{V}}^{(t)}$, and $\text{\metric}^{(t)}$ at each output position $t$, then averaging over positions to obtain a per-sample score. Position-resolved \metric curves $t \mapsto \text{\metric}^{(t)}$ are themselves diagnostic: a model that grounds visually for the first few tokens but drifts into prior-driven generation later in the sequence will exhibit a rising curve, a pattern that is invisible at single-token granularity.

\paragraph{Why we defer it}
Two considerations led us to scope the present paper to single-token answers:
(i) yes/no and multiple-choice benchmarks are the field's standard accuracy-vs-bias evaluation settings, and aligning \metric with them lets us compare it sample by sample against the benchmark's own correctness labels; and
(ii) reporting \metric on free-form generation requires studying different decoding methods and variable length aggregation over batch, each of which is a non-trivial empirical study. We view these as additive scope rather than blockers, and the architectural changes required are limited to the input and output layers.

\subsection{Correctness-Based Value Function}
\label{app:correctness_value}
The value function in \cref{eq:value_function} is label-free. When a ground-truth answer $c^\star$ is available, the game can instead be grounded in correctness by replacing the reference distribution in the link $g$ of \cref{eq:value_function}. Instead of the text-only prediction $p_{\text{LM}}$, we measure divergence from the one-hot distribution $e_{c^\star}$ on the correct option, $g_{\text{gt}}(\cdot) = \mathrm{KL}\left(e_{c^\star} \,\|\, \cdot\right)$. So that a positive value marks an undesirable pull, as in \cref{eq:value_function}, we take the difference in the opposite order:
\begin{equation}
\label{eq:correctness_value}
\begin{aligned}
v_{\text{gt}}(S) \;&=\; \mathrm{KL}\!\left(e_{c^\star} \,\|\, p_S\right) - \mathrm{KL}\!\left(e_{c^\star} \,\|\, p_\emptyset\right) \\
&=\; \log p_\emptyset(c^\star) - \log p_S(c^\star),
\end{aligned}
\end{equation}
where, as in \cref{eq:value_function}, all distributions are renormalized over the $C$ option tokens at the answer position. Again $v_{\text{gt}}(\emptyset) = 0$ by construction. A positive Shapley value $\phi_i^{\text{gt}}$ means player $i$ pulls the model away from the correct answer, and a negative one means it pulls the model toward it. The orientation matches \cref{eq:value_function}, so \metric (\cref{eq:lpb}) applies to $\phi^{\text{gt}}$ unchanged.

\paragraph{Direct visual-grounding diagnosis}
Under \cref{eq:correctness_value}, the per-modality mean $\bar{\phi}^{\text{gt}}_{\mathcal{V}}$ (\cref{eq:mean_shapley}) directly measures visual grounding: it is negative when the image, on average, pulls the prediction toward the correct answer. Since the two games (with and without ground truth labels) share the same link and orientation and differ only in their reference, they are directly comparable, and reading them together for each player separates four cases. A player with $\phi_i > 0$ under \cref{eq:value_function} and $\phi_i^{\text{gt}} > 0$ under \cref{eq:correctness_value} pulls toward the prior and away from the correct answer. A player with $\phi_i < 0$ and $\phi_i^{\text{gt}} < 0$ is grounded evidence that corrects the prior. The two opposite-sign cases arise when the prior and the correct answer agree, where \metric is uninformative by design (\cref{sec:discussion}). On prior-conflict data such as VLind-Bench-LP, where $p_{\text{LM}}$ is typically wrong, we would expect the two games to be closely coupled. On data where the prior is often right, such as POPE and VQAv2, they can diverge, and $v_{\text{gt}}$ could help distinguish prior-following errors from errors in which vision overrides the prior but points to the wrong answer.

\paragraph{Results}
We compute $\phi^{\text{gt}}$ for Qwen3-VL-8B with KernelSHAP ($K{=}512$) on the VLind-Bench-LP and POPE validation splits and score each sample with \metric (\cref{eq:lpb}). \metric under $v_{\text{gt}}$ predicts errors with an AUROC of 0.990 (95\% CI 0.973--1.000) on VLind-Bench-LP and 0.957 (0.932--0.977) on POPE. On VLind-Bench-LP this matches the label-free game (0.991). 

\paragraph{Cost}
The rest of the pipeline is unchanged. The coalition cache stores the $C$ option logits of every sampled coalition and of the grand coalition (\cref{app:coalition-generation}), so $p_S(c^\star)$ is available for all $K$ cached coalitions without new coalition passes. The KernelSHAP target (\cref{eq:phi_star}) and the explainer carry over with $v_{\text{gt}}$ in place of $v$. The price is the label: $v_{\text{gt}}$ cannot be used at deployment, as a label-free fine-tuning objective (\cref{sec:actionable}), or in the prompt search (\cref{app:prompt_refinement}). This is why the main paper uses \cref{eq:value_function}.
\begin{table*}[t]
  \centering
  \small
  \caption{\textbf{Imputer ablation} on Qwen3-VL-4B / POPE ($N{=}100$). KL is $\mathrm{KL}(p_{\text{LM}} \,\|\, p)$, where $p$ is the option distribution with every visual token removed by the imputer and the question kept. Spearman is the rank correlation, across samples, of each imputer's \metric with that of attention masking.}
  \label{tab:imputer}
  \vspace{3mm}
  \begin{tabular}{lccccc}
    \toprule
    \textbf{Imputer} & \textbf{KL} $\downarrow$ & \textbf{\metric} & \textbf{\metric (correct)} & \textbf{\metric (wrong)} & \textbf{Spearman} \\
    \midrule
    \textbf{Attn.\ mask (ours)} & \textbf{0.37} & $-0.330$ & $-0.411$ & $+0.181$ & -- \\
    Mask emb.          & 1.45 & $-0.439$ & $-0.467$ & $+0.034$ & 0.9862 \\
    Mean emb.          & 1.38 & $-0.608$ & $-0.662$ & $-0.154$ & 0.9474 \\
    Patch drop (black) & 2.72 & $-0.570$ & $-0.651$ & $-0.148$ & 0.9862 \\
    Patch drop (gray)  & 3.27 & $-0.426$ & $-0.516$ & $-0.030$ & 0.9714 \\
    Blur region        & 3.14 & $-0.363$ & $-0.422$ & $+0.138$ & 0.9822 \\
    \bottomrule
  \end{tabular}
  \vspace{-3mm}
\end{table*}

\section{Prompt Templates}
\label[appendix]{app:prompts}

Each sample is rendered with the minimal user-visible text required by the answer mode. We add no system or pre-prompt text of our own; any default system message from a model's chat template is held fixed across coalitions, like the rest of the chat-template scaffolding (see \cref{sec:problem_formulation}). Prompts are composed as $\langle\text{question}\rangle + [\langle\text{options}\rangle] + \langle\text{post-prompt}\rangle$. The post-prompt enforces the answer-format expected by the evaluator and is identical across all samples within a dataset. In the rendered prompts below, \texttt{[Image]} marks image-player content; angle-bracketed spans are sample-specific text players.

\paragraph{POPE and VQAv2 (yes/no)} The question is followed by a fixed instruction tail. The question $\mathcal{Q}$ is the only attributable text player.

\begin{promptbox}[title=POPE and VQAv2 (yes/no)]
[Image]\\
<Question> \\
Respond with only Yes or No. Do not include any other text.\\
Answer:
\end{promptbox}

\paragraph{VLind-Bench-LP} VLind-Bench-LP embeds the answer instruction in the question itself. The output vocabulary is \texttt{\{True, False\}} rather than \texttt{\{Yes, No\}}. Only the \texttt{$\langle$statement$\rangle$} span counts as an attributable text player.

\begin{promptbox}[title=VLind-Bench-LP]
[Counterfactual image]\\
Statement: $\langle$statement$\rangle$\\
Based on the image, is the given statement true or false? Forget real-world common sense and just follow the information provided in the image. Only respond in True or False.
\end{promptbox}

\paragraph{General $C$-way multiple-choice} The cooperative-game formulation in \cref{sec:problem_formulation} natively supports any $C$-way multiple-choice answer space, of which the yes/no datasets above are the $C{=}2$ special case. For a $C$-way item, the user-visible string is the question, an enumerated option block (\texttt{A.\,$\langle$option$\rangle$}, \texttt{B.\,$\langle$option$\rangle$}, \ldots), and a fixed instruction tail in which \texttt{$\langle$letters$\rangle$} expands to \texttt{A/B/$\ldots$} according to $C$. The question and the content of each option (not its letter) are the attributable text players; the trailing instruction is held fixed. We include the template here to illustrate that the explainer, value function, and \metric aggregation transfer to general MCQ benchmarks without modification, even though all our experiments use yes/no tasks.

\begin{promptbox}[title=General $C$-way MCQ template]
[Image]\\
$\langle$question$\rangle$\\
Options:\\
A. $\langle$c$_0\rangle$\\
B. $\langle$c$_1\rangle$\\
\quad$\vdots$\\
Respond with only the letter of the correct option ($\langle$letters$\rangle$). Do not include any other text.\\
Answer:
\end{promptbox}

Per-model formatting (system tags, image tokens, role markers) is handled by each model's own chat template. No model-specific rewrites are applied to the user-visible text above.

\section{Implementation Details}
\label[appendix]{app:implementation}

\paragraph{Vision processing} For Qwen-VL, InternVL, and Molmo, whose visual token count grows with input resolution, we repeatedly halve the longer image side until it is at most 160\,px and resize to that size with Lanczos filtering, preserving the aspect ratio. The longer side therefore lies in $(80, 160]$\,px, and the processor sees a single tile. Tarsier resizes to its own fixed processor resolution and is left unchanged. Every \metric score, and every accuracy paired with \metric, is computed on this downscaled input.

\paragraph{Spatial super-tokens} To keep the cooperative game balanced between modalities (\cref{sec:problem_formulation}), we group adjacent visual patches into $K_s{\times}K_s$ super-tokens, each treated as one player. On the patch grid of size $H{\times}W$, $K_s = \max\!\bigl(1, \mathrm{round}(\sqrt{HW/16})\bigr)$, which gives $\lceil H/K_s \rceil \cdot \lceil W/K_s \rceil$ image players. Per-sample \metric ordering is robust to the grouping: re-running KernelSHAP under three grouping conventions spanning $12$ to ${\sim}71$ visual players preserves it at Spearman $0.987$-$0.991$ across all pairs.

\paragraph{KernelSHAP reference} We run KernelSHAP over the same super-token and question-token players, with the attention-mask imputer and the value function of \cref{eq:value_function}. Each sample uses $1{,}500$ coalition evaluations.

\paragraph{\method explainer} The explainer reads the hidden states of layer $18$ of the frozen VLM. Each token's input is the hidden state with $\kappa(\mathbf{x})$ appended, projected to width $512$. The encoder has $6$ Pre-LN blocks with grouped-query attention ($8$ query heads, $2$ key-value heads), SwiGLU feed-forward networks of width $2 \times 512$, RMSNorm, dropout $0.3$, and no positional encoding. The token head reads each encoded token with $\kappa(\mathbf{x})$ appended again. Training uses AdamW (learning rate $3 \times 10^{-4}$, weight decay $0.01$), one-cycle cosine schedule over $250$ epochs with $10\%$ warmup, batch size $16$, gradient clipping at $1.0$, and Huber $\delta = 1$ in standardized \metric units.

The explainer reported in the paper is trained for Qwen3-VL-8B on the VLind-Bench-LP training split together with $9{,}000$ VQAv2 yes/no samples. A random $10\%$ of VLind-Bench-LP is held out for validation and model selection. The coalition cache stores $K = 512$ coalitions per sample (\cref{app:coalition-generation}). KernelSHAP ground truth is computed on $84$ VLind-Bench-LP validation and $64$ POPE samples for evaluation only. \cref{tab:main_results} was run on a cluster of 8 NVIDIA A10 24\,GB GPUs. \cref{tab:efficiency} was measured on a single NVIDIA L40S 48\,GB GPU. Experiments in \cref{sec:actionable} were run on 4 NVIDIA L40S 48\,GB GPUs.
\begin{algorithm*}[t]
\caption{Stage~1: Per-sample Coalition Cache Generation}
\label{alg:coalition-cache}
\begin{algorithmic}[1]
\Require Frozen attention-mask imputer $f$; Sample inputs $(\text{emb}, \text{attn})$, Player mask $\mathcal{P}$, Visual-token mask, Image grid $(T, H, W)$; Option token ids $\mathcal{O} = \{o_1,\dots,o_C\}$; Text-only distribution $p_{\text{LM}}$; Number of coalitions $K$.
\Ensure Cache entry $\bigl(z_{\text{grand}},\, \{M_k\}_{k=1}^{K},\, \{z_k\}_{k=1}^{K},\, g(p_\emptyset),\, p_{\text{LM}}\bigr)$.
\State $M_{\emptyset} \gets \text{attn} \wedge \neg \mathcal{P}$ \Comment{Null: only always-on tokens present}
\State $p_\emptyset \gets \mathrm{softmax}(f(\text{emb}, M_{\emptyset}, \text{attn})[\mathcal{O}])$
\State $z_{\text{grand}} \gets f(\text{emb}, \text{attn}, \text{attn})[\mathcal{O}]$ \Comment{Grand coalition}
\State $(\mathcal{G}, \mathcal{A}) \gets \textsc{BuildSuperTokenGroups}(\text{visual mask}, (T,H,W), \mathcal{P}, \text{attn})$ \\
\Comment{$\mathcal{G}$: $N$ player groups (visual super-tokens $\cup$ question tokens); $\mathcal{A}$: always-on indices}
\If{$N = 0$} \Comment{Degenerate sample with no Shapley players}
    \State \Return $(z_{\text{grand}},\, \{M_{\emptyset}\}^{K},\, \{z_{\text{grand}}\}^{K},\, g(p_\emptyset),\, p_{\text{LM}})$
\EndIf
\For{$k = 1, \ldots, K$}
    \State Sample size $m_k \in \{1,\ldots,N-1\}$ from $\mathrm{Cat}(\pi)$ with $\pi(m) \propto \tfrac{1}{m(N-m)}$
    \State Sample $S_k \in \{0,1\}^{N}$ uniformly at random with $\|S_k\|_1 = m_k$
    \If{$k$ is even} $S_k \gets \mathbf{1} - S_{k-1}$ \Comment{Antithetic pair}
    \EndIf
    \For{each group $g \in \mathcal{G}$ with token positions $\mathcal{T}_g$}
        \State $M_k[\mathcal{T}_g] \gets S_k[g]$ \Comment{Expand group-level masks to token level}
    \EndFor
    \State $M_k[\mathcal{A}] \gets 1$ \Comment{Always-on tokens forced present}
    \State $z_k \gets f\bigl(\text{emb},\, M_k,\, \text{attn}\bigr)[\mathcal{O}]$ \Comment{Coalition pass}
\EndFor
\State \Return $\bigl(z_{\text{grand}},\, \{M_k\},\, \{z_k\},\, g(p_\emptyset),\, p_{\text{LM}}\bigr)$
\end{algorithmic}
\end{algorithm*}

\section{Imputer Ablation}
\label[appendix]{app:imputer}

We compare six imputers on POPE with Qwen3-VL-4B ($N{=}100$ samples, KernelSHAP with $1{,}000$ coalitions over the players of \cref{app:implementation}). Each imputer realizes an absent player differently:
\begin{itemize}[leftmargin=*,itemsep=1pt,topsep=2pt]
  \item \textbf{Attention masking (ours)} zeroes the absent token's attention-mask entry and leaves its embedding unchanged.
  \item \textbf{Mask embedding} replaces absent text and visual tokens with the VLM's learned padding-token embedding.
  \item \textbf{Mean embedding} replaces absent visual tokens with the dataset-mean visual embedding and absent text tokens with the dataset-mean text embedding.
  \item \textbf{Patch drop} replaces absent visual tokens with those of a constant-fill copy of the image, black $(0,0,0)$ or gray $(127,127,127)$.
  \item \textbf{Blur region} replaces absent visual tokens with those of a Gaussian-blurred copy of the image (radius $5$\,px, applied after the resize of \cref{app:implementation}).
\end{itemize}
The two image-bank imputers (patch drop, blur) encode the filled or blurred image once per sample and reuse its visual tokens for every coalition. For absent text players, they fall back to attention masking, as in our imputer. Pixel-space and embedding replacement substitute other signals for the removed evidence (\eg, black pixels still carry information), whereas attention masking excludes the token outright. We omit inpainting-based imputers, whose behavior depends on the inpainting model.

\Cref{tab:imputer} first checks the imputers themselves. Under each imputer we remove every visual token while keeping the question, and measure the KL divergence $\mathrm{KL}(p_{\text{LM}} \,\|\, p)$ from the text-only distribution to the resulting option distribution $p$. A faithful removal of the image should reproduce the text-only pass, so lower is better. The remaining columns report \metric statistics and the Spearman rank correlation of each imputer's \metric with that of attention masking, across the $100$ samples. A sample counts as correct when its full-input prediction matches the label, which does not depend on the imputer. Attention masking comes closest to the text-only pass, though every other imputer ranks samples by \metric much as attention masking does (Spearman $\ge 0.947$). Absolute \metric levels shift with the imputer, but the gap between correct and wrong samples stays similar ($0.49$--$0.59$).
\section{\metric Design Ablations}
\label[appendix]{app:metric_ablations}

We ablate the design choices of \metric (\cref{sec:lpb_metric}). All variants are computed from the same KernelSHAP attributions (\cref{app:implementation}), so only the aggregation changes. $\bar{\phi}_{\mathcal{T}}$ is the sum of the text attributions divided by the number of text players, and $\bar{\phi}_{\mathcal{V}}$ is the sum of the visual attributions divided by the number of visual players. We write $\Phi_{\mathcal{T}} = \sum_{i \in \mathcal{T}} \phi_i$ and $\Phi_{\mathcal{V}} = \sum_{i \in \mathcal{V}} \phi_i$ for the unnormalized modality sums. Unless noted, we experiment on Qwen3-VL-8B and report AUROC for identifying errors on POPE / VLind-Bench-LP, with correctness taken at the same resized input as \metric.

\paragraph{Hard hinge vs.\ softplus} The intended vision term is the hard hinge $\mathrm{ReLU}(-\bar{\phi}_{\mathcal{V}})$, which rewards vision that overrides the prior and neutralizes the rest. Softplus approximates it differentiably, deviating from it by less than $\ln 2$ for either sign of $\bar{\phi}_{\mathcal{V}}$. Across nine models and $842$ VLind-Bench-LP samples, the mean deviation is $0.0090$ ($0.0061$ on the $19.7\%$ of samples with $\bar{\phi}_{\mathcal{V}} > 0$). The two forms agree at Spearman $\rho \geq 0.9965$ across samples for every model ($0.9988$ pooled).

\paragraph{Per-player means vs.\ sums} Replacing the per-player means $\bar{\phi}_{\mathcal{T}}, \bar{\phi}_{\mathcal{V}}$ with the raw modality sums $\Phi_{\mathcal{T}}, \Phi_{\mathcal{V}}$ changes AUROC by at most $0.003$ on either dataset.


\section{Coalition Generation}
\label[appendix]{app:coalition-generation}

\Cref{alg:coalition-cache} details the per-sample procedure used in Stage~1 to build the coalition cache consumed by the \method explainer (\cref{sec:seq2shap}). The frozen VLM is wrapped by an attention-mask imputer $f$. Passing a token mask $M \in \{0,1\}^{L}$ to $f$ realizes the coalition by zeroing the attention-mask entries of absent players, leaving structural (always-on) tokens unchanged.  The player-level coalition $S_k \in \{0,1\}^N$ of the main text is expanded to such a token mask $M_k$ before it is passed to $f$. For each sample, $\texttt{emb}$ holds the VLM's input embeddings, $\texttt{attn}$ its attention mask over the $L$ input tokens. $\mathcal{P} \in \{0,1\}^L$ marks the tokens of attributable players (visual and question tokens), the visual-token mask marks the visual tokens among them, and $(T,H,W)$ is the image's patch grid (frames, height and width).

For each coalition we cache the answer-position logits $z_k$ over the $C$ options $\mathcal{O} = \{o_1, \ldots, o_C\}$.
Per sample, we also cache the null divergence $g(p_\emptyset)$ and the text-only distribution $p_{\text{LM}}$, which comes from a separate pass with the image omitted. Each $z_k$ then gives $v(S_k) = g(p_\emptyset) - g\bigl(\mathrm{softmax}(z_k)\bigr)$ (\cref{eq:value_function}), with $g(\cdot) = \mathrm{KL}(p_{\text{LM}} \,\|\, \cdot)$.

We use $K = 512$. \textsc{BuildSuperTokenGroups} groups visual patches into super-tokens as described in \cref{app:implementation}. Coalitions are drawn at the player level and expanded to tokens, so all tokens of a super-token are present or absent together. Each even-indexed coalition is the complement of the one before it (antithetic pairs, \cref{alg:coalition-cache}), which reduces the variance of the regression target.

\section{Seq2SHAP Ablation}
\label[appendix]{app:seq2shap}

\paragraph{Input layer} \method reads the hidden states of one layer of the frozen VLM. We ablate this choice on Qwen3-VL-8B and Qwen3-VL-4B (both $D{=}36$ decoder layers) with layers $9$, $18$ and $36$ (the final hidden state). Every other setting follows \cref{app:implementation}, and each configuration is trained with two seeds. All explainers use the training data of the main explainer (VLind-Bench-LP and the yes/no subset of VQAv2). We score them on the $84$ VLind-Bench-LP validation samples and, on the $64$ out-of-distribution POPE test samples, against the KernelSHAP-$1{,}500$ attributions of \cref{app:implementation}. We report the Pearson correlation of per-sample \metric ($r_{\text{\metric}}$). \Cref{tab:layer_ablation} summarizes the results.

\begin{table}[t]
  \centering
  \footnotesize
  \caption{\textbf{Input-layer ablation} for the \method explainer, trained on VLind-Bench-LP and VQAv2. We report $r_{\text{\metric}}$ against KernelSHAP-$1{,}500$ on the VLind-Bench-LP validation split (at the selected epoch) and on the POPE test split. Values are means over two seeds. Best per model in bold.}
  \label{tab:layer_ablation}
  \setlength{\tabcolsep}{6pt}
  \begin{tabular}{llcc}
    \toprule
    \textbf{Model} & \textbf{Layer} & \textbf{VLind-Bench-LP} $\uparrow$ & \textbf{POPE} $\uparrow$ \\
    \midrule
    Qwen3-VL-8B & $9$  & $0.985$          & $0.983$          \\
                & $18$ & $\mathbf{0.991}$ & $\mathbf{0.989}$ \\
                & $36$ & $0.990$          & $0.980$          \\
    \midrule
    Qwen3-VL-4B & $9$  & $0.958$          & $0.978$          \\
                & $18$ & $\mathbf{0.979}$ & $0.988$          \\
                & $36$ & $0.975$          & $\mathbf{0.989}$ \\
    \bottomrule
  \end{tabular}
  \vspace{-3mm}
\end{table}

\paragraph{Results} Layer $18$ is the best or near-best input in every column. On Qwen3-VL-8B it is best on both evaluation sets ($0.991$ on VLind-Bench-LP and $0.989$ on POPE). On Qwen3-VL-4B it is best on VLind-Bench-LP ($0.979$) and within $0.001$ of layer $36$ on POPE. Layer $9$ is the weakest input on VLind-Bench-LP for both models, most clearly on Qwen3-VL-4B ($0.958$ vs.\ $0.979$ for layer $18$). Layer $36$ is the weakest on POPE for Qwen3-VL-8B ($0.980$ vs.\ $0.989$), so neither end of the network is a reliable choice. We therefore keep the middle layer.

\section{Offline Training Target and Its Equivalence to FastSHAP}
\label[appendix]{app:offline}

\paragraph{Setup}
Fix a sample $\mathbf{x}$ with $N = |\mathcal{N}|$ players and its $K$ cached coalitions $S_1, \ldots, S_K \in \{0,1\}^{N}$, drawn from the Shapley kernel $\pi$ (\cref{sec:seq2shap}, Stage~1). Let $b_k = v(S_k) - v(\emptyset)$, $A = \frac{1}{K}\sum_k S_k S_k^\top$, $\mathbf{c} = \frac{1}{K}\sum_k b_k S_k$, and $\Delta = v(\mathcal{N}) - v(\emptyset)$. On the cache $\{S_k\}_{k=1}^K$, the FastSHAP objective (\cref{eq:fastshap}) for a prediction $\boldsymbol{\phi} \in \mathbb{R}^{N}$ is the Monte Carlo estimate
\begin{equation}
\begin{aligned}
\mathcal{L}_{\text{fs}}(\boldsymbol{\phi}) 
&\;=\; \frac{1}{K}\sum_{k=1}^{K} \bigl(b_k - S_k^\top \boldsymbol{\phi}\bigr)^2 \\
&\;=\; \boldsymbol{\phi}^\top A \boldsymbol{\phi} - 2\,\mathbf{c}^\top \boldsymbol{\phi} + \tfrac{1}{K}\textstyle\sum_k b_k^2 .
\end{aligned}
\end{equation}

\paragraph{Proposition}
Let $\boldsymbol{\phi}^{*}$ minimize $\mathcal{L}_{\text{fs}}$ subject to $\mathbf{1}^\top \boldsymbol{\phi} = \Delta$ (\cref{eq:phi_star}). Then for every $\boldsymbol{\phi}$ with $\mathbf{1}^\top \boldsymbol{\phi} = \Delta$,
\begin{equation}
\mathcal{L}_{\text{fs}}(\boldsymbol{\phi}) \;=\; \mathcal{L}_{\text{fs}}(\boldsymbol{\phi}^{*}) + (\boldsymbol{\phi} - \boldsymbol{\phi}^{*})^\top A\, (\boldsymbol{\phi} - \boldsymbol{\phi}^{*}).
\end{equation}

\paragraph{Proof}
Expand $\mathcal{L}_{\text{fs}}$ around $\boldsymbol{\phi}^{*}$ with $\mathbf{d} = \boldsymbol{\phi} - \boldsymbol{\phi}^{*}$:
$\mathcal{L}_{\text{fs}}(\boldsymbol{\phi}) = \mathcal{L}_{\text{fs}}(\boldsymbol{\phi}^{*}) + 2\,\mathbf{d}^\top (A\boldsymbol{\phi}^{*} - \mathbf{c}) + \mathbf{d}^\top A\, \mathbf{d}$.
The KKT condition of the constrained problem gives $A\boldsymbol{\phi}^{*} - \mathbf{c} = \lambda \mathbf{1}$ for a scalar multiplier $\lambda$. Thus the cross term is $2\lambda\, \mathbf{1}^\top \mathbf{d} = 2\lambda(\mathbf{1}^\top\boldsymbol{\phi} - \Delta) = 0$. \hfill$\square$

The constant $\mathcal{L}_{\text{fs}}(\boldsymbol{\phi}^{*})$ does not depend on $\theta$. Thus, on efficient predictions, minimizing FastSHAP on the cache equals minimizing an $A$-weighted distance to $\boldsymbol{\phi}^{*}$, and $\boldsymbol{\phi}^{*}$ is a sufficient statistic of the $K$ cached coalitions. In practice, we solve \cref{eq:phi_star} in closed form with a ridge of $10^{-6}$ on $A$.

\paragraph{From the target to the loss}
The Proposition shows that $\boldsymbol{\phi}^{*}$ carries all the coalition information. It does not require us to regress on $\boldsymbol{\phi}^{*}$ with the $A$-weighted norm. We train with \cref{eq:training_loss} instead. The per-modality cosine matches the direction of $\boldsymbol{\phi}^{*}$ within each modality, and the Huber term on the standardized \metric matches its scale through the quantity the metric uses. Both terms reach their minimum at $\hat{\boldsymbol{\phi}} = \boldsymbol{\phi}^{*}$.

\paragraph{Cost}
Online FastSHAP evaluates the VLM on fresh coalitions at every training step. With $E$ epochs and $K$ coalitions per sample per epoch, it needs $E \cdot K$ VLM forward passes per sample, and the VLM must stay in memory during training. Offline training needs $K$ VLM passes per sample once (Stage~1), plus one full pass for the hidden states $\mathbf{h}$. The explainer then trains on $(\mathbf{h}, \kappa, \boldsymbol{\phi}^{*})$ alone, with no VLM in memory. With our settings ($K = 512$, $E = 250$), the offline scheme needs about $90\times$ fewer VLM passes. Each training step costs only an explainer forward and backward pass, so we can tune the explainer many times on one cache.

\section{Shapley Axiom Checks}
\label[appendix]{app:axioms}

\method is an amortized estimator, so we test efficiency, dummy and symmetry empirically. We use the Qwen3-VL-8B explainer of \cref{app:implementation} on the first $32$ VLind-Bench-LP validation samples ($802$ players), which are held out from its training. For each sample we draw $32$ base coalitions $S$ over the players, with a uniform size and then a uniform subset, and evaluate $v(S)$ and $v(S \cup \{i\})$ for every player $i \notin S$ under \cref{eq:value_function}. Coalition values are unbounded, so thresholds are expressed as a fraction of the magnitude of the explained total, $|\Delta| = |v(\mathcal{N}) - v(\emptyset)|$ (\cref{eq:phi_star}). As a reference, we run the same tests on the KernelSHAP attributions ($1{,}500$ coalitions, \cref{app:implementation}) of the same samples. Attribution gaps are reported relative to the within-sample spread, the standard deviation of the attributions over a sample's players. These tests are approximate rather than a full verification that \method satisfies the axioms. We only use $32$ base coalitions per sample, alongside re-evaluated coalition values and tolerance-based definitions of near-dummy players, which adds measurement noise. This is why KernelSHAP, which satisfies the axioms in expectation, shows non-zero gaps in every test, and why \method can score below it on symmetry. \Cref{tab:axioms} summarizes the results.

\begin{table}[t]
  \centering
  \footnotesize
  \caption{\textbf{Shapley axiom checks} on $32$ held-out VLind-Bench-LP validation samples ($802$ players) for Qwen3-VL-8B. Efficiency is the mean gap $|\sum_i \phi_i - \Delta| / |\Delta|$. Dummy is the mean $|\phi_i|$ of near-dummy players, and symmetry the mean attribution difference within near-exchangeable pairs, both in within-sample spreads (lower is better). Tolerances are $5\%$ / $10\%$ of $|\Delta|$, which give $5$ / $22$ near-dummy players and $38$ / $239$ pairs.}
  \label{tab:axioms}
  \setlength{\tabcolsep}{4pt}
  \begin{tabular}{lccc}
    \toprule
    \textbf{Method} & \textbf{Efficiency} $\downarrow$ & \textbf{Dummy} $\downarrow$ & \textbf{Symmetry} $\downarrow$ \\
    \midrule
    KernelSHAP-1{,}500 & $0.11$                & $0.17$ / $0.29$ & $0.41$ / $0.40$ \\
    \method            & $1.2 \times 10^{-7}$ & $0.20$ / $0.28$ & $0.19$ / $0.24$ \\
    \bottomrule
  \end{tabular}
  \vspace{-3mm}
\end{table}

\paragraph{Efficiency} The training target $\boldsymbol{\phi}^{*}$ satisfies efficiency exactly: \cref{eq:phi_star} constrains $\sum_{i \in \mathcal{N}} \phi^{*}_i = \Delta$, and we solve it in closed form with a ridge of $10^{-6}$ on $A$. The explainer enforces it on its own predictions through the additive shift of \cref{sec:seq2shap}, applied both in training and at inference. The remaining gap $|\sum_i \hat{\phi}_i - \Delta| / |\Delta|$ is $1.2 \times 10^{-7}$ on average (at most $4.7 \times 10^{-7}$), which is floating-point precision. KernelSHAP's gap against our re-evaluated $v(\mathcal{N})$ is $0.11$ on average (at most $0.51$).

\paragraph{Dummy} We mark player $i$ as a near-dummy if $\max_S |v(S \cup \{i\}) - v(S)| / |\Delta|$ is below a tolerance. At tolerances of $5\%$ and $10\%$ ($5$ and $22$ players), the explainer assigns these players $|\hat{\phi}_i| = 0.20$ and $0.28$ within-sample spreads, close to KernelSHAP's $0.17$ and $0.29$. The explainer therefore drives near-dummy players toward zero about as well as KernelSHAP.

\paragraph{Symmetry} From the same marginal contributions, we identify near-exchangeable pairs whose contributions agree within the tolerance on every shared base coalition, with at least $4$ shared. At $5\%$ ($38$ pairs) the explainer's attributions within a pair differ by $0.19$ within-sample spreads, against $0.41$ for KernelSHAP. At $10\%$ ($239$ pairs) the explainer differs by $0.24$ and KernelSHAP by $0.40$, and the same holds for pairs with above-median contributions ($0.21$ vs.\ $0.42$).

\section{More Qualitative Results}
\label[appendix]{app:heatmaps}

\begin{figure*}[p]
  \centering
  \includegraphics[width=\linewidth]{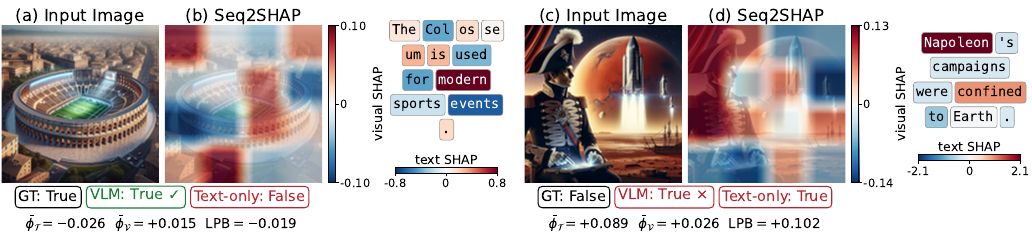}\\[1mm]
  \includegraphics[width=\linewidth]{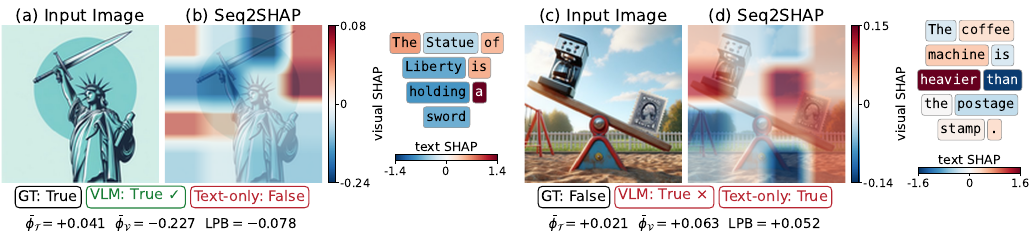}\\[1mm]
  \includegraphics[width=\linewidth]{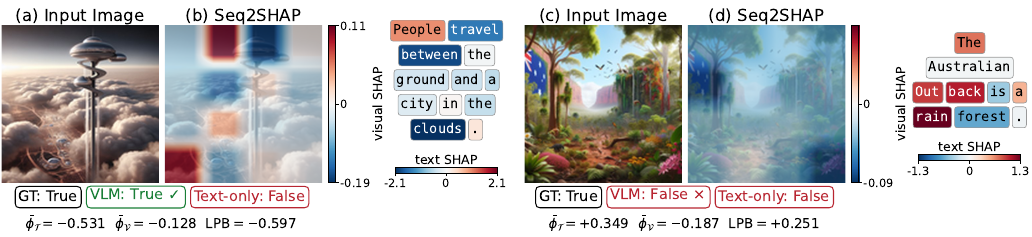}\\[1mm]
  \includegraphics[width=\linewidth]{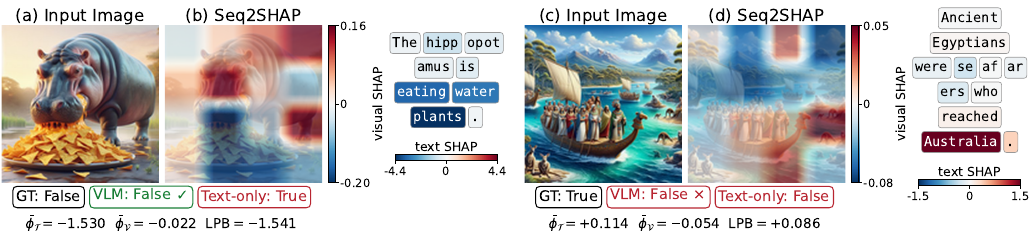}\\[1mm]
  \includegraphics[width=\linewidth]{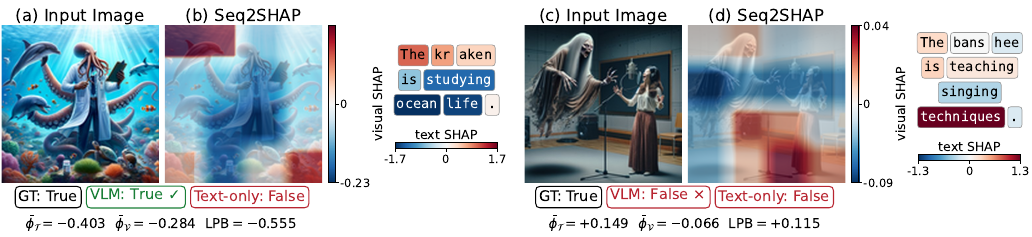}
  \caption{\textbf{Token-level \method attribution on ten VLind-Bench-LP samples from Qwen3-VL-8B.} Left: five correctly answered samples with negative \metric. Right: five incorrectly answered samples with positive \metric. Samples are drawn at random within each group. Red pushes the prediction toward the text-only prediction $p_{\text{LM}}$, and blue pushes it away.}
  \label{fig:appx_heatmaps_vlind}
\end{figure*}

\begin{figure*}[p]
  \centering
  \includegraphics[width=\linewidth]{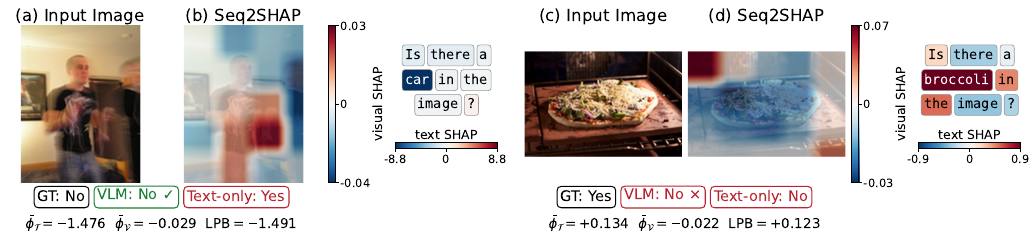}\\[1mm]
  \includegraphics[width=\linewidth]{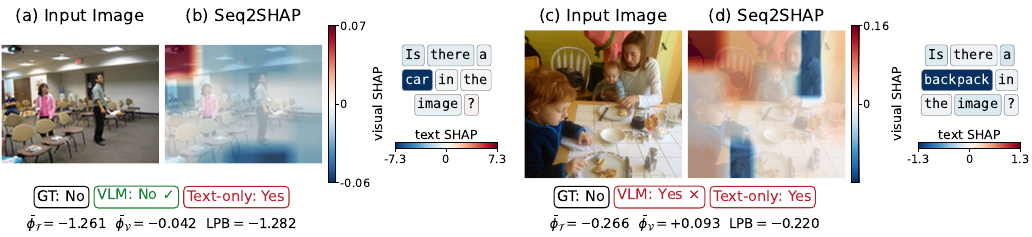}\\[1mm]
  \includegraphics[width=\linewidth]{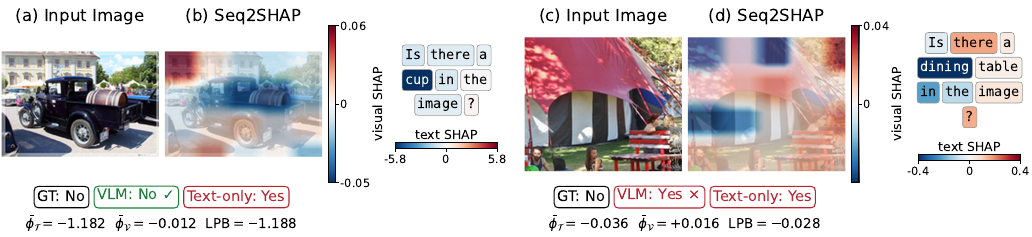}\\[1mm]
  \includegraphics[width=\linewidth]{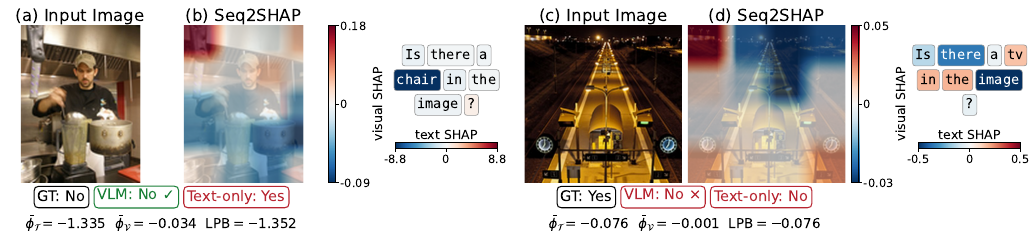}\\[1mm]
  \includegraphics[width=\linewidth]{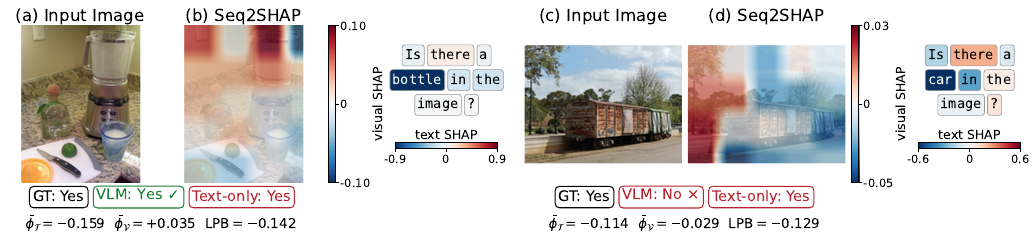}
  \caption{\textbf{Token-level \method attribution on ten POPE samples from Qwen3-VL-8B.} Left: five correctly answered samples. Right: five incorrectly answered samples. Samples are drawn at random within each group. POPE is never seen by the explainer during training. Red pushes the prediction toward the text-only prediction $p_{\text{LM}}$, and blue pushes it away.}
  \label{fig:appx_heatmaps_pope}
\end{figure*}

\Cref{fig:appx_heatmaps_vlind,fig:appx_heatmaps_pope} extend \cref{fig:shap_heatmaps} to ten more samples each from VLind-Bench-LP and POPE, using the final Qwen3-VL-8B \method explainer (\cref{app:implementation}) with the efficiency normalization of \cref{sec:seq2shap} applied, and the same rendering. On each dataset we draw five correctly and five incorrectly answered samples at random (seed $0$), excluding the two samples of \cref{fig:shap_heatmaps}. On VLind-Bench-LP we draw correct samples only among those with negative \metric and incorrect samples only among those with positive \metric. Each row pairs a correct sample (left) with an incorrect one (right). Token shading shows per-token text attributions and the image overlay shows per-region visual attributions. Red pushes the prediction toward the text-only prediction $p_{\text{LM}}$, and blue pushes it away. Each panel reports $\bar{\phi}_{\mathcal{T}}$, $\bar{\phi}_{\mathcal{V}}$, and \metric (\cref{eq:mean_shapley,eq:lpb}). The chips under each image give the ground truth, the VLM prediction, and the text-only prediction.

\section{\metric Fine-Tuning}
\label[appendix]{app:finetune}

\paragraph{Training loss} For each training sample we run four prefill passes under the attention-mask imputer: all players present ($p_{\mathcal{N}}$), visual players absent ($p_{\mathcal{T}} = p_{\text{LM}}$), question-text players absent ($p_{\mathcal{V}}$), and all players absent ($p_\emptyset$). Each pass is restricted to the yes/no option logits. With $v(S) = \mathrm{KL}(p_{\text{LM}} \,\|\, p_\emptyset) - \mathrm{KL}(p_{\text{LM}} \,\|\, p_S)$, the exact two-player Shapley values are
\begin{equation}
\begin{aligned}
\psi_{\mathcal{T}} &= \tfrac{1}{2}\bigl[v(\mathcal{T}) + v(\mathcal{N}) - v(\mathcal{V})\bigr], \\
\psi_{\mathcal{V}} &= \tfrac{1}{2}\bigl[v(\mathcal{V}) + v(\mathcal{N}) - v(\mathcal{T})\bigr].
\end{aligned}
\end{equation}
We divide $\psi_{\mathcal{T}}$ by the number of question tokens and $\psi_{\mathcal{V}}$ by the number of visual super-tokens (\cref{app:implementation}), and apply \cref{eq:lpb}. The training loss is $\mathcal{L} = \mathcal{L}_{\text{CE}} + \lambda\,\mathcal{L}_\text{\metric}$ with $\lambda{=}10$, where $\mathcal{L}_{\text{CE}}$ is the cross-entropy of $p_{\mathcal{N}}$ against the ground-truth yes/no answer.

\paragraph{Training} We LoRA-fine-tune each VLM on $6{,}000$ samples drawn at random from the yes/no subset of VQAv2~\cite{goyal2017making}. LoRA (rank $16$, $\alpha{=}32$, no dropout, no bias) adapts the attention and feed-forward projections of every language-model decoder layer ($q$, $k$, $v$, $o$, gate, up, and down projections). The vision encoder, token embeddings, and output head stay frozen. We train for $400$ optimizer steps with AdamW (learning rate $3{\times}10^{-5}$, no schedule), one sample per micro-batch with $16$ accumulation steps (an effective batch size of $16$), and gradient clipping at $1.0$. Images are resized as in \cref{app:implementation}. The control arm uses the same recipe with $\lambda{=}0$. Both arms are trained with the same three random seeds.

\paragraph{Evaluation} We compare three states per model: the base VLM, the control, and the \metric fine-tune. Accuracy is measured on all VLind-Bench-LP samples. \metric is computed with KernelSHAP ($1{,}000$ coalitions) over the textual and visual players of \cref{app:implementation}, on the first $200$ VLind-Bench-LP samples by index, the same subset for every state. Each state recomputes its own text-only distribution $p_{\text{LM}}$ from a text-only pass of that state.

\paragraph{Ablation study on $\lambda$} We sweep $\lambda \in \{0.3, 1, 3, 10, 30\}$ on all eight models with the same three seeds as \cref{tab:finetune} (\cref{tab:lambda_ablation}). Relative to the same-seed control, mean VLind-Bench-LP accuracy change is positive at every value and grows with $\lambda$: $+0.9$, $+0.9$, $+2.0$, $+4.5$, and $+5.3$ points. We use $\lambda{=}10$ for training stability.

\paragraph{Seed variability} Across the three seeds, the standard deviation of the accuracy change of \emph{+\metric} relative to the same-seed control (\cref{tab:finetune}) is $2.0$, $1.5$, $1.4$, $1.1$, $1.7$, $2.2$, $2.1$, and $3.8$ points for Qwen2-VL-7B, Qwen2.5-VL-7B, Qwen3-VL-2B, Qwen3-VL-4B, Qwen3-VL-8B, Tarsier-7B, InternVL3-8B, and Molmo2-8B, respectively.

\begin{table}[t]
\centering
\footnotesize
\caption{\textbf{Ablation on $\lambda$.} Change in VLind-Bench-LP accuracy (points) relative to the same-seed control for each $\lambda$, averaged over three seeds.}
\label{tab:lambda_ablation}
\begin{tabular}{lccccc}
\toprule
& \multicolumn{5}{c}{$\lambda$} \\
\cmidrule(lr){2-6}
\textbf{Model} & $0.3$ & $1$ & $3$ & $10$ & $30$ \\
\midrule
Qwen2-VL-7B & $+0.1$ & $+0.2$ & $+0.9$ & $+7.2$ & $+8.2$ \\
Qwen2.5-VL-7B & $-2.8$ & $-3.1$ & $-3.0$ & $-0.4$ & $+0.8$ \\
Qwen3-VL-2B & $+1.4$ & $+2.1$ & $+3.1$ & $+7.9$ & $+5.8$ \\
Qwen3-VL-4B & $+3.2$ & $+2.8$ & $+4.6$ & $+6.0$ & $+9.7$ \\
Qwen3-VL-8B & $+2.7$ & $+3.4$ & $+4.8$ & $+6.3$ & $+9.2$ \\
Tarsier-7B & $+0.4$ & $-0.6$ & $+2.0$ & $+3.1$ & $+4.0$ \\
InternVL3-8B & $+0.6$ & $+1.0$ & $+0.9$ & $+3.4$ & $+4.0$ \\
Molmo2-8B & $+1.3$ & $+1.8$ & $+2.5$ & $+2.2$ & $+0.5$ \\
\midrule
Mean & $+0.9$ & $+0.9$ & $+2.0$ & $+4.5$ & $+5.3$ \\
\bottomrule
\end{tabular}
\end{table}
\section{LPB-Guided Prompt Refinement}
\label[appendix]{app:prompt_refinement}

Besides fine-tuning (\cref{sec:actionable}), \metric can be optimized at deployment time without changing model weights. For each sample we run a small genetic search over instruction prefixes prepended to the question, with \metric as fitness (lower is better). An LLM mutator (\texttt{gpt-5.4-mini}~\cite{openai2026gpt54mini}) proposes new prefixes. A \emph{seed-only} control scores the seed prefix with no search.

The search runs over the failure pool, the samples the model answers incorrectly with no prefix. We run it on Qwen3-VL-8B~\cite{bai2025qwen3vl} and VLind-Bench-LP, using the \method explainer trained for this model (\cref{app:implementation}) and the KL value function (\cref{eq:value_function}). The seed prefix is ``\texttt{Please answer the following:}''. The failure pool holds the $374$ samples answered incorrectly at the resized input (\cref{app:implementation}). LPB-guided search flips $\mathbf{113}$ ($\mathbf{30.2\%}$) to correct vs.\ $\mathbf{7}$ ($\mathbf{1.9\%}$) for the seed-only control. 
The discovered prefixes are short, generic visual-grounding instructions (\cref{tab:ga_flip_examples}). 

\label{app:prompt_refinement:config}

The genetic search uses population $P{=}8$, $G{=}5$ generations, and elitism $E{=}2$ (the top-$E$ by \metric carry over unchanged). The initial population holds the empty prefix, the seed prefix, and $6$ mutator variants of the seed ($3$ exploration and $3$ exploitation). Each generation the mutator produces $P-E=6$ new prefixes: $4$ exploitation candidates, which refine the lowest-\metric prefixes so far, and $2$ exploration candidates, which must use a different framing from everything in the history, even at higher \metric. The mutator sees the all-time top-$8$ and bottom-$3$ \metric scorers, so it can avoid re-proposing known-bad regions of the prefix space. Total per-sample budget is $P + G\cdot(P-E) = 38$ \metric evaluations and $1 + G = 6$ mutator calls. The search returns the candidate with the lowest \metric over all generations, and a sample counts as flipped if the model answers it correctly under that prefix.

Each candidate is scored by the trained \method explainer, not by KernelSHAP. For each candidate we rebuild the prompt with the prefix, run the full-input VLM pass for the hidden states and the prediction, recompute $p_{\text{LM}}$ with a text-only pass on the prefixed prompt, and apply \cref{eq:lpb} to the explainer's per-modality means. Prefix tokens are held fixed and are not players, so $\bar{\phi}_{\mathcal{T}}$ averages only over the original question tokens. $\bar{\phi}_{\mathcal{V}}$ averages over the visual super-tokens (\cref{app:implementation}).


\subsection{Mutator Prompts}
\label{app:prompt_refinement:prompts}

The mutator is invoked once per generation with the system message and user-message template below. It sees only candidate prefixes and their scalar \metric scores. It is never shown the image, question, correct label, predicted label, or option probabilities, so any downstream accuracy gain must propagate through \metric.

\begin{promptbox}[title=Mutator system message (verbatim)]
You are helping interpretability researchers craft instruction prefixes that encourage a vision-language model to ground its answers in the image rather than relying on language priors (text-only guesses). Each prefix is scored by a diagnostic metric called LPB. A LOWER score indicates the model attended more to the image and less to text-only shortcuts, which is the desired behavior. Your job is to propose new natural-English prefix variants based on the supplied history of past trials (top performers and bottom performers across all generations so far). Each request specifies how many EXPLOIT variants (refine the lowest-LPB pattern) and how many EXPLORE variants (deliberately different framing/strategy from anything in history, even at the cost of higher LPB) to produce. Emit them in that order. Output a JSON object with a single key `prefixes' whose value is a list of strings (no markdown, no code fences). Each prefix should be plain English, under 200 characters, and end with a trailing space.
\end{promptbox}

\begin{promptbox}[title=Mutator user-message template]
Generate $\langle N\rangle$ new prefix variants. Of those:\\
\hspace*{1em}- $\langle N_{\text{exploit}}\rangle$ EXPLOIT variants: refine the strategy of the lowest-LPB entries below, pushing further in that direction.\\
\hspace*{1em}- $\langle N_{\text{explore}}\rangle$ EXPLORE variants: pursue substantively different framings (different verbs, focus, or persona) from anything in history, even at the cost of higher LPB. Avoid restating the top entries' strategy.\\[2pt]
TOP (lowest-LPB) trials so far: $\langle$JSON list of \{prompt, lpb\} for the top-$8$ history$\rangle$\\[2pt]
BOTTOM (highest-LPB) trials so far (do not regenerate these patterns): $\langle$JSON list of \{prompt, lpb\} for the bottom-$3$ history$\rangle$\\[2pt]
Emit the $\langle N_{\text{exploit}}\rangle$ EXPLOIT variants first, then the $\langle N_{\text{explore}}\rangle$ EXPLORE variants. Return a JSON list of $\langle N\rangle$ strings.
\end{promptbox}

\subsection{Discovered Flip-Inducing Prefixes}
\label{app:prompt_refinement:examples}

\Cref{tab:ga_flip_examples} lists the $10$ flip-inducing prefixes with the largest \metric reduction ($\Delta\metric$, most negative first). They are drawn from the $106$ samples of the failure pool ($n{=}374$) that the search flips to correct and the seed-only control does not. For each sample we report the probability of the predicted class under the empty prefix and under the discovered prefix; the post-search prediction is bolded and matches the correct label in every row.

\begin{table*}[t]
\centering
\footnotesize
\caption{Top-$10$ flip-inducing prefixes by $\Delta\metric$ on Qwen3-VL-8B / VLind-Bench-LP, from the $106$ flips that the seed-only control does not recover (failure pool $n{=}374$). ``Pred prob empty $\to$ searched'' shows the probability mass on the originally-predicted class under the empty prefix and on the post-search prediction (bolded); in all $10$ rows the post-search prediction matches the ground-truth label. Ground-truth labels follow the counterfactual image, not common sense. All ten rows flip \emph{True} to a correct \emph{False} (VLind-Bench-LP's counter-commonsense framing primes \emph{True}); across all $113$ flips, $40$ go from \emph{False} to \emph{True}, so the search is direction-agnostic.}
\label{tab:ga_flip_examples}
\vspace{3mm}
\setlength{\tabcolsep}{4pt}
\renewcommand{\arraystretch}{1.2}
\begin{tabular}{c p{0.27\linewidth} c c p{0.36\linewidth}}
\toprule
\textbf{idx} & \textbf{Statement (GT)} & \shortstack{\textbf{Pred prob}\\\textbf{empty $\to$ searched}} & $\boldsymbol{\Delta\metric}$ & \textbf{Discovered prefix} \\
\midrule
483 & Drones were not used in the Revolutionary War. (F) & T $1.00 \to$ \textbf{F $1.00$} & $-1.378$ & \emph{``From a new angle, narrate the scene like a witness describing what is present:''} \\
717 & The elephants are terrestrial animals. (F) & T $0.73 \to$ \textbf{F $1.00$} & $-1.249$ & \emph{``Adopt an evidence-checker mindset and answer from the image alone:''} \\
127 & Ancient Egyptians did not have access to modern construction technology. (F) & T $1.00 \to$ \textbf{F $1.00$} & $-1.098$ & \emph{``State only the details you can directly see in the image:''} \\
197 & Online banking apps were not available to medieval merchants. (F) & T $1.00 \to$ \textbf{F $1.00$} & $-1.086$ & \emph{``Play the image detective and list the clues you can directly see:''} \\
331 & Spartans did not have access to modern technology or social media. (F) & T $1.00 \to$ \textbf{F $1.00$} & $-1.032$ & \emph{``From a visual standpoint, explain the scene as if you were seeing it for the first time:''} \\
461 & Napoleon's campaigns were confined to Earth. (F) & T $1.00 \to$ \textbf{F $1.00$} & $-1.023$ & \emph{``From the standpoint of a witness, describe what the image reveals before answering:''} \\
523 & The washing machine is heavier than the rubber duck. (F) & T $0.88 \to$ \textbf{F $1.00$} & $-0.900$ & \emph{``As a careful analyst, infer nothing beyond the picture and state only what is shown:''} \\
51  & The Mount Everest base camp is a starting point for climbers. (F) & T $0.56 \to$ \textbf{F $1.00$} & $-0.858$ & \emph{``As an image analyst, cite only visible details before answering:''} \\
175 & Laptops were not used by medieval monks for manuscript writing. (F) & T $0.90 \to$ \textbf{F $1.00$} & $-0.834$ & \emph{``Respond only with evidence you can directly see in the image:''} \\
381 & Smartphones were not available for medieval kings to receive news updates. (F) & T $0.98 \to$ \textbf{F $1.00$} & $-0.801$ & \emph{``Study the picture carefully and use only visible details in your response:''} \\
\bottomrule
\end{tabular}
\end{table*}

\end{document}